%% file: 0_main_combined.tex
\PassOptionsToPackage{dvipsnames,table,xcdraw}{xcolor}
\documentclass[10pt,twocolumn,letterpaper]{article}

\usepackage{iccv}              
\usepackage[accsupp]{axessibility} 

\input{preamble}

\definecolor{iccvblue}{rgb}{0.21,0.49,0.74}

\def\paperID{12205} 
\def\confName{ICCV}
\def\confYear{2025}
\title{FlowEdit: Inversion-Free Text-Based Editing Using Pre-Trained Flow Models}

\author{
  \begin{minipage}[t]{\textwidth}
    \centering
    Vladimir Kulikov \hskip 0.5cm
    Matan Kleiner \hskip 0.5cm
    Inbar Huberman-Spiegelglas  \hskip 0.5cm
    Tomer Michaeli \\
    \vspace{0.1cm}
    Technion – Israel Institute of Technology\\
  \end{minipage}
}

\begin{document}
\twocolumn[{%
\renewcommand\twocolumn[1][]{#1}%
\maketitle
\begin{center}
    \centering
    \captionsetup{type=figure}
    \includegraphics[width=\textwidth]{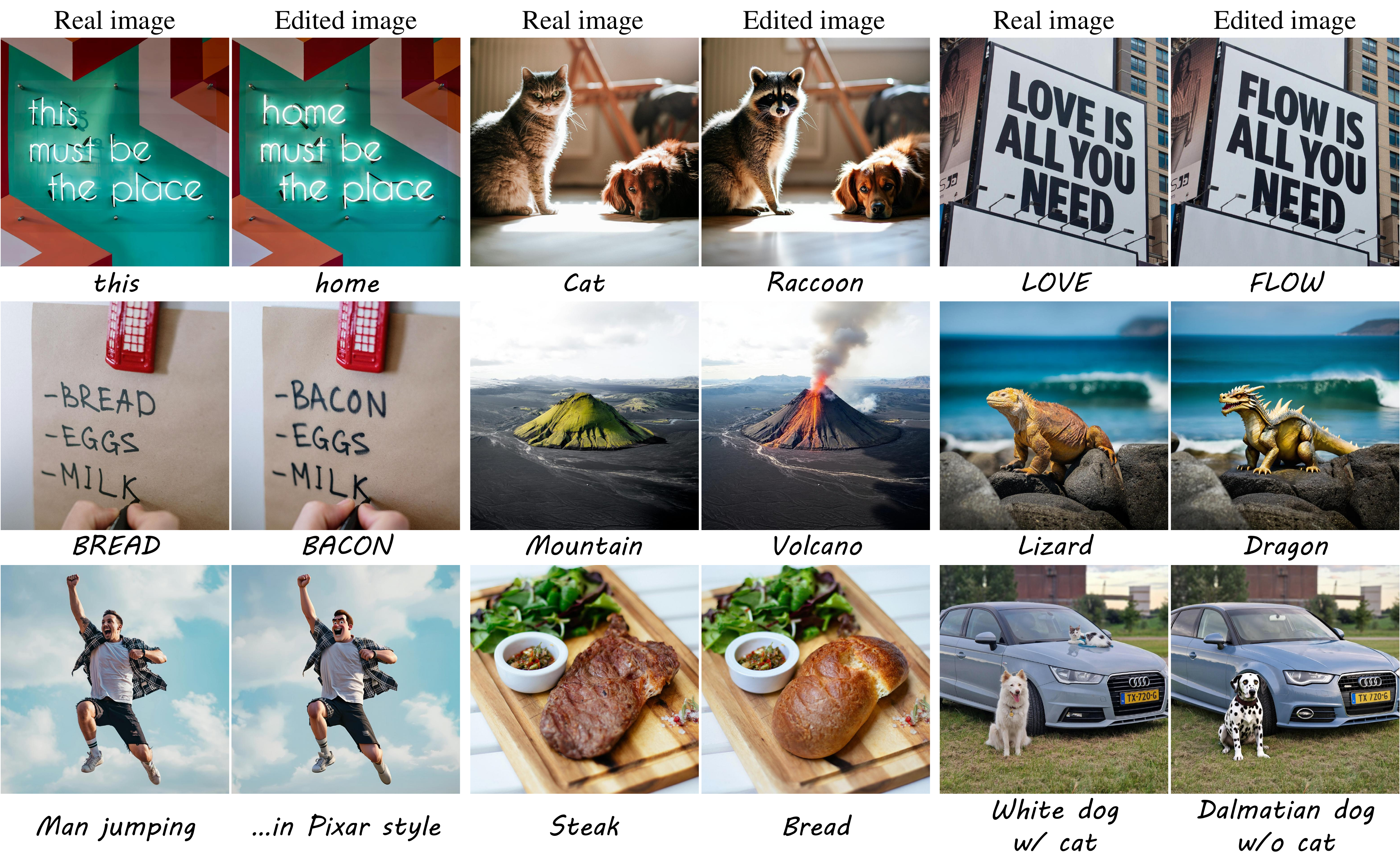}
    \captionof{figure}{\textbf{\ours{}.} We present an inversion-free, optimization-free and model agnostic method for text-based image editing using pre-trained flow models. As opposed to the editing-by-inversion paradigm, \ours{} constructs an ODE that directly maps the source image distribution to the target image distribution (corresponding to the source and target prompts). This ODE achieves a lower transport cost and thus leads to better structure preservation, achieving state of the art results on complex editing tasks. From left to right, top to bottom, the first five images were obtained with FLUX and the rest with Stable Diffusion~3. Text indicates changes in the prompts.}
    \label{fig:teaser}
\end{center}%
}]

\input{sec/0_abstract}    
\input{sec/1_intro}
\input{sec/2_related_work}
\input{sec/3_preliminaries}
\input{sec/4_method}

\input{sec/5_evaluation}

\input{sec/6_conclusion}

\clearpage

{
    \small
    \bibliographystyle{ieeenat_fullname}
    \bibliography{
    main}
}

\clearpage

\onecolumn

\renewcommand\thefigure{S\arabic{figure}}    
\setcounter{figure}{0}  
\renewcommand{\thesection}{\Alph{section}}
\setcounter{section}{0}
\renewcommand{\thetable}{S\arabic{table}}
\setcounter{table}{0}
\renewcommand{\theequation}{S\arabic{equation}}
\setcounter{equation}{0}
\renewcommand{\thealgorithm}{S\arabic{algorithm}}
\setcounter{algorithm}{0}

\input{sec/X_suppl}


\end{document}

%% file: preamble.tex
\usepackage{mathtools}

\newcommand{\ours}[1]{FlowEdit}
\newcommand{\baseurl}{https://matankleiner.github.io/flowedit/}
\newcommand{\webpage}{\href{\baseurl}{webpage}}

\usepackage{algorithm}
\usepackage{algorithmic}
\usepackage[dvipsnames,table,xcdraw]{xcolor}

\usepackage{makecell}  
\usepackage{nicematrix}  
\usepackage[most]{tcolorbox}  
\usepackage{xr}
\usepackage{xr-hyper}
\usepackage{hyperref}

\hypersetup{
	hidelinks,
	colorlinks=true,
	linkcolor=Blue,
	filecolor=Blue,
	citecolor=Blue,
	urlcolor=Blue,
	breaklinks=false
}

\definecolor{tabfirst}{rgb}{1, 0.7, 0.7} 
\definecolor{tabsecond}{rgb}{1, 0.85, 0.7} 
\definecolor{tabthird}{rgb}{1, 1, 0.7} 

\usepackage{tikz}
\usetikzlibrary{decorations.pathreplacing,calc}
\newcommand{\tikzmark}[1]{\tikz[overlay,remember picture] \node (#1) {};}

\newcommand*{\AddNote}[4]{%
    \begin{tikzpicture}[overlay, remember picture]
        \draw [decoration={brace,amplitude=0.8em},decorate, thick,blue]
            ($(#3)!(#1.north)!($(#3)-(0,1)$)$) --  
            ($(#3)!(#2.south)!($(#3)-(0,1)$)$)
                node [align=center, text width=2.5cm, pos=0.5, anchor=west] {#4};
    \end{tikzpicture}
}%

\makeatletter
\newcommand*{\addFileDependency}[1]{
	\typeout{(#1)}
	\@addtofilelist{#1}
	\IfFileExists{#1}{\typeout{File #1 O.K.}}{\typeout{No file #1.}}
}
\makeatother


%% file: sec/0_abstract.tex
\begin{abstract}
Editing real images using a pre-trained text-to-image (T2I) diffusion/flow model often involves inverting the image into its corresponding noise map. 
However, inversion by itself is typically insufficient for obtaining satisfactory results, and therefore many methods additionally intervene in the sampling process. Such methods achieve improved results but are not seamlessly transferable between model architectures. 
Here, we introduce \ours{}, a text-based editing method for pre-trained T2I flow models, which is inversion-free, optimization-free and model agnostic. Our method constructs an ODE that directly maps between the source and target distributions (corresponding to the source and target text prompts) and achieves a lower transport cost than the inversion approach. 
This leads to state-of-the-art results, as we illustrate with Stable Diffusion~3 and FLUX. 
Code and examples are available on the project’s \webpage{}.
\end{abstract}

%% file: sec/1_intro.tex
\section{Introduction}
\label{intro}
Diffusion and flow models generate samples through an iterative process, which starts from pure noise and employs denoising-like steps. Beyond data generation, many methods employ pre-trained diffusion/flow models for editing real signals~\citep{meng2021sdedit,kawar2023imagic,huberman2024edit,tumanyan2023plug, hertz2022prompt, manor2024zero, wu2023tune, qi2023fatezero, cohen24slicedit, haque2023instruct,brack2024leditspp,hertz2023delta,kwon2023diffusion, haas2024discovering}. 
While some techniques rely on test-time optimization~\cite{hertz2023delta,kawar2023imagic}, many methods aim for optimization-free approaches. A common first step in the latter category of methods is \emph{inversion}, \ie extracting the initial noise vector that presumably generates the signal one wishes to edit. This noise vector is then used to generate the edited signal by injecting a different condition to the model, such as a modified text prompt~\citep{wallace2023edict, wu2023latent, huberman2024edit, mokady2023null}. 

In the context of image editing, many works noted that this editing-by-inversion paradigm often leads to insufficient fidelity to the source image (see \eg~\cite{huberman2024edit}). To overcome this, some methods attempt to reduce the errors incurred in the inversion process~\citep{mokady2023null, garibi2024renoise}. However, this rarely eliminates the problem as even when editing generated images, for which the initial noise is precisely known, editing-by-inversion commonly fails to preserve structure. Other methods intervene in the sampling process by injecting internal model representations (\eg attention maps) recorded during the inversion process~\cite{hertz2022prompt, tumanyan2023plug, cao2023masactrl}. These methods achieve improved fidelity but are not easily transferable across model architectures and sampling schemes.

In this work, we present \ours{}, a text-based editing method for pre-trained text-to-image (T2I) flow models, which breaks away from the editing-by-inversion paradigm. \ours{}  is optimization-free and model agnostic, making it easily transferable between models. Rather than mapping the image to noise and the noise back to an edited image, it constructs a direct path between the source and target distributions (corresponding to the source and target text prompts). This path is shorter than that achieved by inversion, and thus maintains better fidelity to the source image. 
We evaluate \ours{} on diverse edits using FLUX~\citep{flux} and Stable Diffusion~3 (SD3)~\citep{esser2024scaling}, and show that it achieves state-of-the-art results across the board (see Fig.~\ref{fig:teaser}).

Our method is motivated by a novel reinterpretation of editing-by-inversion as a direct path between the source and target distributions. Armed with this observation, we derive an alternative direct path that has a lower transport cost, as we show on synthetic examples. 

%% file: sec/2_related_work.tex
\section{Related work}
\label{related}

\begin{figure*}
    \centering
    \includegraphics[width=\textwidth]{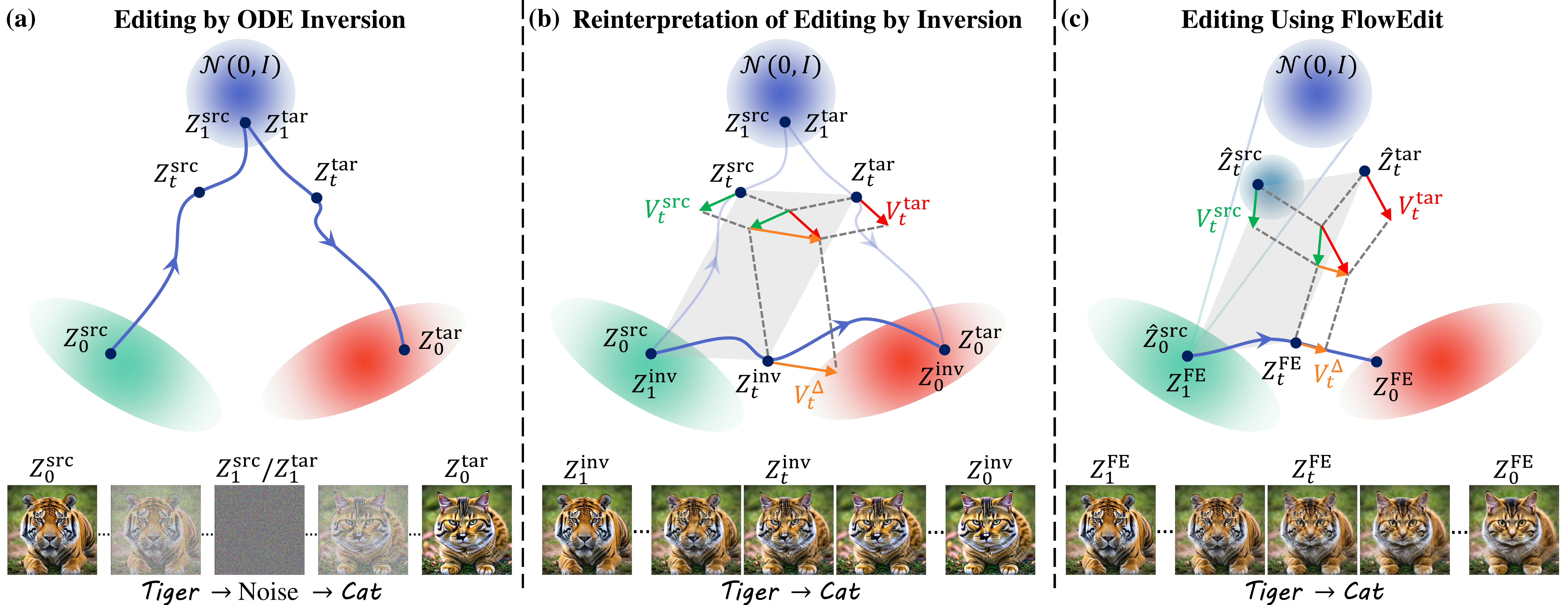}
    \caption{\textbf{Editing by inversion vs.~\ours{}.} \textbf{(a)} In inversion based editing, the source image $Z_0^{\text{src}}$ is first mapped to the noise space by solving the forward ODE conditioned on the source prompt (left path). Then, the extracted noise is used to solve the reverse ODE conditioned on the target prompt to obtain $Z_0^{\text{tar}}$ (right path). The images at the bottom visualize this transition. 
    \textbf{(b)} We reinterpret inversion as a direct path between the source and target distributions (bottom path). 
    This is done by using the velocities calculated during the inversion and sampling (green and red arrows) to calculate an editing direction (orange arrow) that drives the evolution of the direct path $Z_t^{\text{inv}}$ through an ODE.
    The resulting path is noise-free, as demonstrated by the images at the bottom. 
    \textbf{(c)} \ours{} traverses a \emph{shorter} direct path, $Z_t^{\text{FE}}$, without relying on inversion. 
    At each timestep, we directly add random noise to $\hat{Z}_0^{\text{src}}$ to obtain $\hat{Z}_t^{\text{src}}$ and use that direction to create $\hat{Z}_t^{\text{tar}}$ from $Z_t^{\text{FE}}$(gray parallelogram). We then calculate the corresponding velocities and average over multiple realizations (not shown in the figure) to obtain the next ODE step (orange arrow).
    The images at the bottom demonstrate our noise-free path. }
    \label{fig:stoch_interp}
\end{figure*}


To edit a real image using a pre-trained diffusion model, some methods use optimization over the image itself~\citep{hertz2023delta, nam2024contrastive, koo2024posterior, dream2024jeongsol}. These methods utilize the generative prior of a T2I diffusion model as a loss, which they optimize to push the image to comply with the user-provided prompt. Recently,~\citet{yang2024text} suggested a similar method for flow models. These methods are resource intensive at test-time.   

Many zero-shot editing methods do not require optimization. The first step of most of these methods is image-to-noise inversion~\citep{tumanyan2023plug, hertz2022prompt, huberman2024edit, cao2023masactrl, brack2024leditspp, tsaban2023ledits, wallace2023edict, deutch2024turboedit, wu2023latent, samuel2023regularized}. However, the noise map obtained through naive inversion~\cite{song2020ddim, wu2023latent} is generally unsuitable for effective editing~\cite{huberman2024edit, hertz2022prompt, mokady2023null}. This is often attributed to the inaccurate inversion process, leading to attempts to improve the inversion accuracy~\cite{mokady2023null, han2023improving, wallace2023edict, miyake2023negative}. Yet, even exact inversion on synthetic data leads to unsatisfactory editing~\citep{huberman2024edit}. To address this, many methods extract during the inversion stage structural information implicitly encoded within the model architecture. They then inject it in the sampling process to achieve better structure preservation~\cite{hertz2022prompt, tumanyan2023plug, cao2023masactrl, parmar2023zero, brack2024leditspp}.   
These methods are typically tailored for specific model architectures and sampling methods, limiting their transferability to new settings, such as to flow models~\cite{flux, esser2024scaling}. 
Several recent methods were proposed for editing with flow models \citep{rout2024rfinversion, wang2024taming, avrahami2024stable}. However, these methods rely on inversion and in some cases also on feature injection, and thus suffer from the same limitations as their diffusion-model counterparts.

Unlike previous works, our method does not rely on inversion. It maps from source to target distributions without traversing through the Gaussian distribution. It also avoids optimization and does not intervene in the model internals, making it easily adaptable to new models.

%% file: sec/3_preliminaries.tex
\section{Preliminaries}
\label{pre}

\subsection{Rectified Flow models}
Generative flow models attempt to construct a transportation between the distributions of two random vectors, $X_0$ and $X_1$, defined by an ordinary differential equation (ODE) over time $t\in[0,1]$,
\begin{equation}\label{eq:flow_orig}
    dZ_t=V(Z_t, t) \,dt.
\end{equation}
Here, $V$ is a time-dependent velocity field, usually parameterized by a neural network, which has the property that if the boundary condition at $t=1$ is $Z_1=X_1$,  then $Z_0$ is distributed like $X_0$. It is common to choose $X_1\sim\mathcal{N}(0,\boldsymbol{I})$, which allows to easily draw samples from the distribution of $X_0$. This is done by initializing the ODE at $t=1$ with a sample of a standard Gaussian vector, and numerically solving the ODE backwards in time to obtain a sample $Z_0$ from the distribution of $X_0$ at time $t=0$.

Rectified flows~\citep{albergo2023building, lipman2023flow, liu2023flow} are a particular type of flow models, which are trained such that the marginal at time $t$ corresponds to a linear interpolation between $X_0$ and $X_1$, 
\begin{equation}
\label{eq:linear_interpolant}
    Z_t \sim (1-t) X_0 + t X_1. 
\end{equation} 
Rectified flows have the desired property that their sampling paths are relatively straight. This allows using a small number of discretization steps when solving the ODE.

Text-to-image flow models employ a velocity field $V(X_t, t, C)$ that depends on a text prompt $C$. Such models are trained on pairs of text and image data, $(C,X_0)$, so as to allow sampling from the conditional distribution of $X_0|C$.

\subsection{Image editing using ODE inversion}
\label{sec:ode_inversion}
Suppose we are given a real image, $X^{\text{src}}$, which we want to edit by providing a text prompt $c_{\text{src}}$ describing the image and a text prompt $c_{\text{tar}}$ describing the desired edit. A common approach to do so is by using a pre-trained text-conditioned diffusion/flow model and employing \emph{inversion}. Specifically, let us denote the text conditioned velocities as $V^{\text{src}}(X_t, t)\triangleq V(X_t, t, c_{\text{src}})$ and $V^{\text{tar}}(X_t, t)\triangleq V(X_t, t, c_{\text{tar}})$. 
Methods relying on inversion start by extracting the initial noise map corresponding to the source image. This is done by traversing the forward process defined by the ODE
\begin{equation}
\label{eq:inv_edit_fwd}
    dZ^{\text{src}}_t= V^{\text{src}}(Z^{\text{src}}_t,t)\,dt,
\end{equation}
starting at $t=0$ from the source image, $Z^{\text{src}}_0=X^{\text{src}}$, and reaching a noise map $Z^{\text{src}}_1$ at $t=1$. 
The process $Z^{\text{src}}_t$ is the path shown on the left of Fig.~\ref{fig:stoch_interp}a. Then, the reverse ODE,
\begin{equation}
\label{eq:inv_edit_rev}
    dZ^{\text{tar}}_t = V^{\text{tar}}(Z^{\text{tar}}_t,t)\,dt,
\end{equation}
is solved backward in time, starting from $t=1$ with the extracted noise, $Z^{\text{tar}}_1=Z_1^{\text{src}}$, and reaching an edited image $Z^{\text{tar}}_0$ at $t=0$. This is the path shown on the right of Fig.~\ref{fig:stoch_interp}a.

Editing by inversion tends to produce unsatisfactory results, and is thus usually accompanied by feature map injection. While the sub-optimal results are commonly blamed on the inaccurate discretization of the ODE, they are also encountered when editing generated images and using their ground-truth initial noise maps \cite{huberman2024edit}. 
Our method retains the simplicity of editing by inversion and achieves state of the art results without any intervention in the model internals.

%% file: sec/4_method.tex
\section{Reinterpretation of editing by inversion}
\label{reinterpretation}
Before presenting \ours{}, we reinterpret the editing-by-inversion approach. This will serve to motivate our method.

Inversion based editing transports between the source and target distributions, while passing through the distribution of Gaussian noise. 
However, this approach can also be expressed as a direct path between the source and target distributions. Namely, we can construct a path $Z^{\text{inv}}_t$ that starts at $t=1$ in the source distribution and reaches the target distribution at $t=0$. 
Specifically, given the forward and reverse flow trajectories, $Z^{\text{src}}_t$ and $Z^{\text{tar}}_t$, 
we define
\begin{equation}
\label{eq:direct_coupling}
    Z^{\text{inv}}_t=Z^{\text{src}}_0+Z^{\text{tar}}_t-Z^{\text{src}}_t.
\end{equation}
Note that when going from $t=1$ to $t=0$ we transition from the source image, $Z^{\text{inv}}_1=Z^{\text{src}}_0=X^{\text{src}}$ (because $Z^{\text{tar}}_1=Z^{\text{src}}_1$), to the edited image, $Z^{\text{inv}}_0=Z^{\text{tar}}_0$. 
This path is depicted at the bottom of Fig.~\ref{fig:stoch_interp}b, in which the gray parallelogram visualizes the relation \eqref{eq:direct_coupling}. 

Let us express \eqref{eq:direct_coupling} as an ODE. Differentiating both sides of~\eqref{eq:direct_coupling} and substituting \eqref{eq:inv_edit_fwd} and \eqref{eq:inv_edit_rev}, this relation becomes
\begin{equation}
\label{eq:direct_coupling_ODE}
    dZ^{\text{inv}}_t= V^{\Delta}_{t}(Z^{\text{src}}_{t},Z^{\text{tar}}_{t})\,dt,
\end{equation}
where $V^{\Delta}_{t}(Z^{\text{src}}_{t},Z^{\text{tar}}_{t}) = V^{\text{tar}}(Z^{\text{tar}}_{t},t)-V^{\text{src}}(Z^{\text{src}}_{t},t)$. This is visualized by the green, red, and orange arrows in Fig.~\ref{fig:stoch_interp}b. Now, isolating $Z^{\text{tar}}_{t}$ from \eqref{eq:direct_coupling} and substituting it into \eqref{eq:direct_coupling_ODE}, we get that the direct path is the solution of the ODE
\begin{equation}
\label{eq:direct_coupling_ODE_combined}
    dZ^{\text{inv}}_t= V^{\Delta}_{t}(Z^{\text{src}}_{t},Z^{\text{inv}}_t+Z^{\text{src}}_t-Z^{\text{src}}_0)\,dt 
\end{equation}
with boundary condition $Z^{\text{inv}}_1=Z^{\text{src}}_0$ at $t=1$.

\begin{figure}
    \centering
    \includegraphics[width=\linewidth]{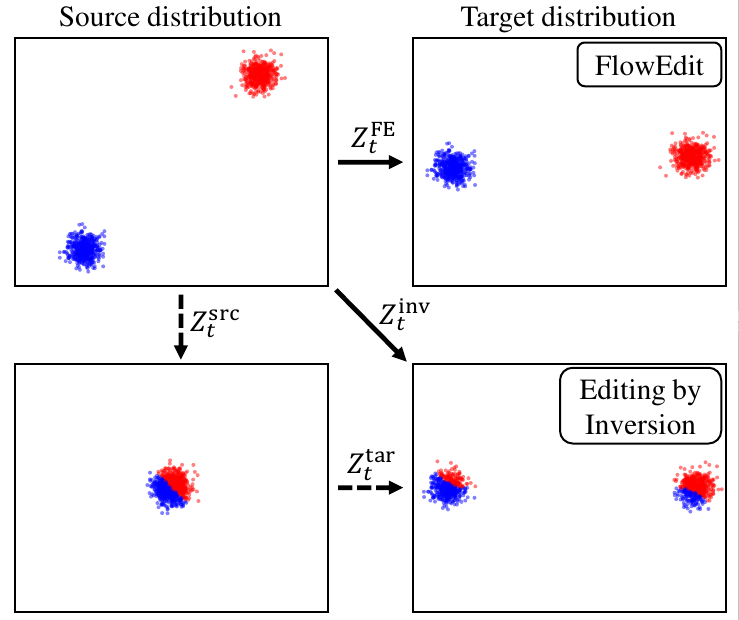}
    \caption{\textbf{Source to target pairings for editing-by-inversion and \ours{}.} Samples from the two modes in the source distribution are colored in blue and red (top left). As can be seen in the right panes, \ours{} results in \emph{separated} blue and red modes, as opposed to editing-by-inversion, where the modes are \emph{intermixed}, indicating higher transport cost in terms of MSE. This can be explained by the Gaussian in the bottom left, through which the editing-by-inversion path must traverse before reaching the target. Please see the supplementary video for animated visualizations.}
    \label{fig:transport2d}
\end{figure}

There exist, of course, many paths connecting  $Z^{\text{src}}_0$ and~$Z^{\text{tar}}_0$. What is so special about the path defined by \eqref{eq:direct_coupling_ODE_combined}? It turns out that images along this path are noise-free, as shown at the bottom of Fig.~\ref{fig:stoch_interp}b. The reason this happens is that the noisy images $Z^{\text{tar}}_t$ and $Z^{\text{src}}_t$ contain roughly the same noise constituent. Therefore, the vectors $V^{\text{tar}}(Z^{\text{tar}}_{t},t)$ and $V^{\text{src}}(Z^{\text{src}}_{t},t)$, which point at the source and target distributions, respectively, remove roughly the same noise component, so that the difference vector $V^{\Delta}_{t}(Z^{\text{src}}_{t},Z^{\text{tar}}_{t})$ (orange arrow in Fig.~\ref{fig:stoch_interp}b), encompasses the difference only between the clean image predictions.

How do images evolve along this direct path? At the beginning of the path ($t$ close to $1$), the noise level in $Z^{\text{src}}_{t}$ and $Z^{\text{tar}}_{t}$ is substantial, and therefore the vector $V^{\Delta}_{t}(Z^{\text{src}}_{t},Z^{\text{tar}}_{t})$ captures only differences in coarse image structures. As~$t$ gets smaller, the noise level drops and higher frequency contents are unveiled. In other words, the path \eqref{eq:direct_coupling_ODE_combined} constitutes a sort of autoregressive coarse-to-fine evolution from $Z^{\text{src}}_0$ to $Z^{\text{tar}}_0$. This can be seen in the images at the bottom of Fig.~\ref{fig:stoch_interp}b, where the first features that get modified are the coarse structures, and the last features to get updated are the fine textures. See App.~\ref{sm:v_delta} for additional illustrations.

\section{\ours{}}
\label{sec:flowedit}

The fact that editing-by-inversion can be expressed as a direct ODE does not imply that it induces a desirable pairing between source and target samples. Figure~\ref{fig:transport2d} shows a simple example, where the source and target distributions are Gaussian mixtures with modes centered at $\{(-\tfrac{15}{\sqrt{2}},-\frac{15}{\sqrt{2}}),(\tfrac{15}{\sqrt{2}},\tfrac{15}{\sqrt{2}})\}$ and $\{(-15,0),(0,15)\}$, respectively. A good editing method should map each mode in the source distribution to its closest mode in the target distribution. In the context of image editing, this means that source images are modified as little as possible, while transporting to the target distribution. However, as seen on the bottom right of Fig.~\ref{fig:transport2d}, this is not what happens in inversion. The reason the induced pairings look the way they look can be understood by inspecting how samples are mapped to their ``initial'' Gaussian noise component (bottom left).

Our goal is to construct an alternative mapping, which leads to lower distances between the source and target samples. Our method, which we coin \emph{\ours{}}, is illustrated on the top right
of Fig.~\ref{fig:transport2d}. As can be seen, it maps each mode in the source distribution to its nearest mode in the target distribution. Compared to editing-by-inversion, \ours{} achieves less than twice as low a transportation cost, measured by the average squared distance between source samples and their paired target samples. We emphasize that our method is based on a heuristic, which is not guaranteed to precisely map to the target distribution. In particular, it does not necessarily coincide with the optimal transport mapping. Yet, as we will see in the context of images, it achieves state-of-the-art results in terms of structure preservation (\ie small transportation cost) while adhering to the target text prompt (generating samples in the target distribution).

How can we depart from the pairings dictated by inversion? Our key idea is to use multiple random pairings, and average their associated velocity fields. More concretely, we propose to use the same ODE as \eqref{eq:direct_coupling_ODE_combined}, but to replace the inversion path $Z^{\text{inv}}_t$ by a different random process with the same marginals. Then, at each timestep, we average the directions corresponding to different realizations of that process. 
Specifically, consider the alternative forward process
\begin{equation}
\label{eq:our_marginals}
    \hat{Z}^{\text{src}}_{t} =(1-t)Z_0^{\text{src}}+tN_{t},
\end{equation}
where $N_{t}$ is a Gaussian process that is statistically independent of $Z^{\text{src}}_0$ and has marginals $N_{t}\sim\mathcal{N}(0,\,1)$ for every $t\in[0,1]$. We construct a path $Z^{\text{FE}}_t$ by solving the ODE
\begin{equation}
\label{eq:our_ODE}
    dZ^{\text{FE}}_t= \mathbb{E}\left[V^{\Delta}_{t}(\hat{Z}^{\text{src}}_{t},Z^{\text{FE}}_t+\hat{Z}^{\text{src}}_t-Z^{\text{src}}_0) \middle| Z^{\text{src}}_0\right]\,dt 
\end{equation}
with boundary condition $Z^{\text{FE}}_1=Z^{\text{src}}_0$ at $t=1$. Note that the expectation here is w.r.t.~$\hat{Z}^{\text{src}}_{t}$ (equivalently $N_t$). Also note that we need not specify the covariance function of $N_t$ as ~\eqref{eq:our_ODE} depends only on the marginals of the process $\hat{Z}^{\text{src}}_{t}$ and not on its entire distribution law.

\begin{algorithm}[tb]
   \caption{Simplified algorithm for \ours{}}
   \label{alg:simple}
\begin{algorithmic}
   \STATE {\bfseries Input:} 
   real image $X^{\text{src}},\big\{t_i\big\}^T_{i=0}, n_{\text{max}}, n_{\text{avg}}$ 
   \STATE {\bfseries Output:} edited image $X^{\text{tar}}$
   \STATE {\bfseries Init:} $Z^{\text{FE}}_{t_{\text{max}}}=X^{\text{src}}_0$
   
   \FOR{$i=n_{\text{max}}$ {\bfseries to} $1$}
   \STATE $N_{t_i} \sim \mathcal{N}(0,\,1)$\tikzmark{top}
   \STATE $Z^{\text{src}}_{t_i} \leftarrow (1-t_i)X^{\text{src}} + t_iN_{t_i}$
   \STATE $Z^{\text{tar}}_{t_i} \leftarrow Z^{\text{FE}}_{t_i} +Z^{\text{src}}_{t_i}- X^{\text{src}}$
   \STATE $V^{\Delta}_{t_i} \leftarrow V^{\text{tar}}(Z^{\text{tar}}_{t_i},t_i)-V^{\text{src}}(Z^{\text{src}}_{t_i},t_i)$\tikzmark{right}\tikzmark{bottom}
   \STATE $Z^{\text{FE}}_{t_{i-1}} \leftarrow Z^{\text{FE}}_{t_i} + (t_{i-1}-t_{i})V^{\Delta}_{t_i}$
   \ENDFOR
   \STATE {\bfseries Return:} $Z^{\text{FE}}_{0} = X_{0}^{\text{tar}}$
\end{algorithmic}
\AddNote{top}{bottom}{right}{Optionally average $n_{\text{avg}}$ samples}
\end{algorithm}

A schematic illustration is shown in Fig.~\ref{fig:stoch_interp}c. The point $\hat{Z}^{\text{src}}_{t}$ on the left corresponds to a single draw of $N_t$. The distribution of $\hat{Z}^{\text{src}}_{t}$ at time $t$ is shown in cyan. For each such draw, we calculate $\hat{Z}^{\text{tar}}_t=Z^{\text{FE}}_t+\hat{Z}^{\text{src}}_t-Z^{\text{src}}_0$, as in the editing-by-inversion ODE. This is indicated by the gray parallelogram.  Now, given $\hat{Z}^{\text{src}}_{t}$ and $\hat{Z}^{\text{tar}}_t$, we compute the velocity field $V^\Delta_t(\hat{Z}^{\text{src}}_{t},\hat{Z}^{\text{tar}}_t)= V^{\text{tar}}(\hat{Z}^{\text{tar}}_t,t)-V^{\text{src}}(\hat{Z}^{\text{src}}_t,t)$ to obtain an update direction. This is illustrated by the green, red, and orange arrows. We repeat this multiple times and average the resulting orange arrows (not shown in the figure) to obtain the final update direction. 

\subsection{Practical considerations}
\label{sec:pratical}

In practice, a discrete set of timesteps $\{t_i\}^{T}_{i=0}$ is used to drive the editing process, where $T$ is the number of discretization steps. Additionally, the expectation in \eqref{eq:our_ODE} is approximated by averaging $n_{\text{avg}}$ model predictions at each timestep. As opposed to the theoretical expectation operator, when taking $n_{\text{avg}}$ to be small, the covariance function of $N_t$ starts playing a role. To obtain a good approximation, it is possible to exploit the averaging that naturally occurs across timesteps. We do so by choosing the covariance function of $N_t$ to satisfy $\mathbb{E}[N_tN_s]=0$ for every $|t-s|>\delta$, where $\delta$ is the ODE discretization step, so that the noise becomes independent across timesteps.

Note that the averaged velocity term in \eqref{eq:our_ODE} corresponds to the difference
$\mathbb{E}[V^{\text{tar}}(\hat{Z}^{\text{tar}}_t,t) | Z^{\text{src}}_0] - \mathbb{E}[V^{\text{src}}(\hat{Z}^{\text{src}}_t,t) | Z^{\text{src}}_0]$, which could be computed by sampling $\hat{Z}^{\text{tar}}_t$ and $\hat{Z}^{\text{src}}_t$ independently. However, we choose to compute them with the same noise instances, which further improves robustness to small values of $n_{\text{avg}}$. 
This is aligned with the observation of \citet{hertz2023delta} in the context of diffusion models, that calculating a difference between correlated noisy marginals reduces artifacts. See Sec.~\ref{subsec:relation_optim} for further discussion.

Similarly to \cite{huberman2024edit,meng2021sdedit}, we define an integer that determines the starting timestep for the process $0 \leq n_{\text{max}} \leq T$, meaning that the process is initialized with $\smash{Z^{\text{FE}}_{t_{n_{\text{max}}}}=X^{\text{src}}}$. When $n_{\text{max}}=T$, the full edit path is traversed and the strongest edit is obtained. When $n_{\text{max}}<T$, the first $(T - n_{\text{max}})$ timesteps are skipped, effectively shortening the edit path. This is equivalent to inversion, where weaker edits are obtained by inverting up to timestep $n_{\text{max}}$, and sampling from there. We illustrate the effect of $n_{\text{max}}$ in App.~\ref{sm:n_max}.

Algorithm~\ref{alg:simple} provides a simplified overview of our algorithm. For a more detailed version, please see Alg.~\ref{alg:full}.

\begin{figure}
    \centering
    \includegraphics[width=\linewidth]{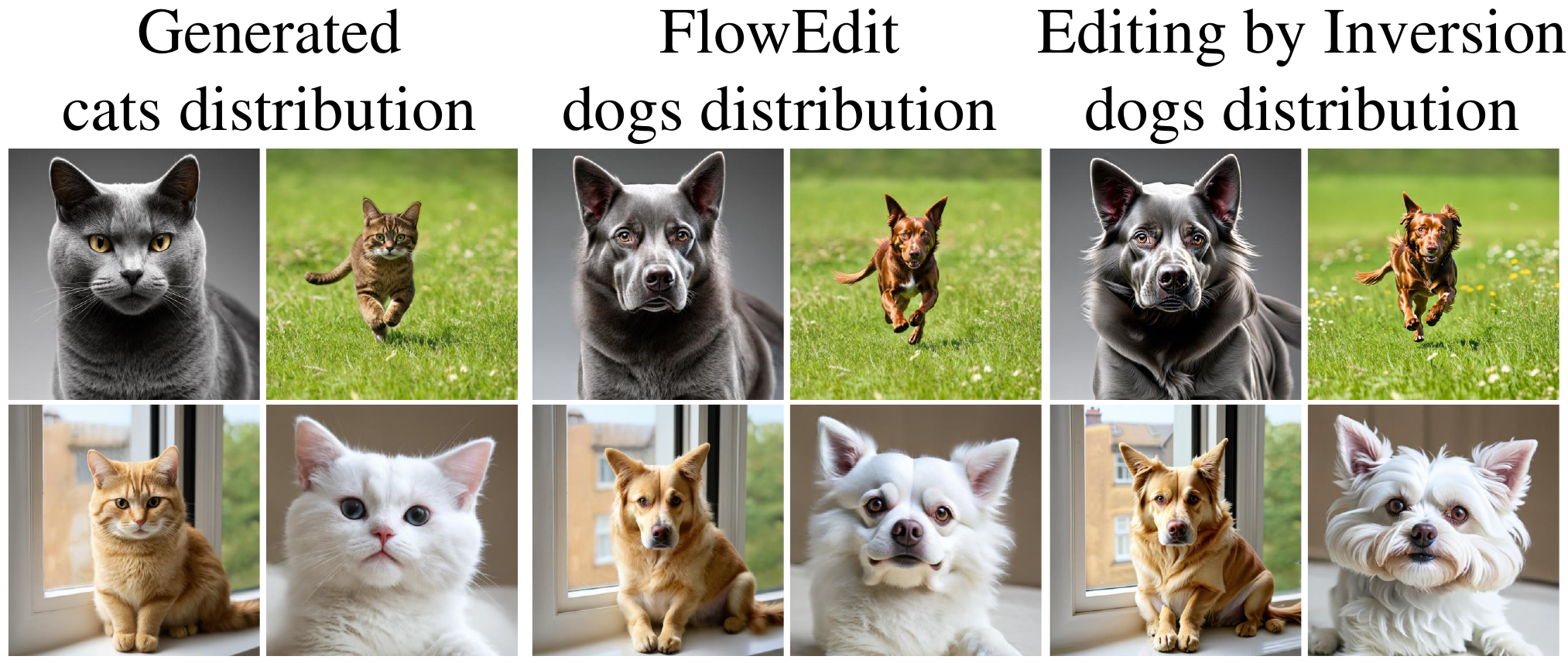}
    \caption{\textbf{Cats and dogs experiment.} We generated 1000 cat images using diverse prompts and edited them to dog images using both \ours{} and inversion. \ours{} outperforms editing by inversion, achieving a lower transport costs and better FID and KID scores (computed against 1000 generated dog images).}
    \label{fig:cats_dogs}
\end{figure}

\begin{figure*}
    \centering
    \includegraphics[width=0.97\textwidth]{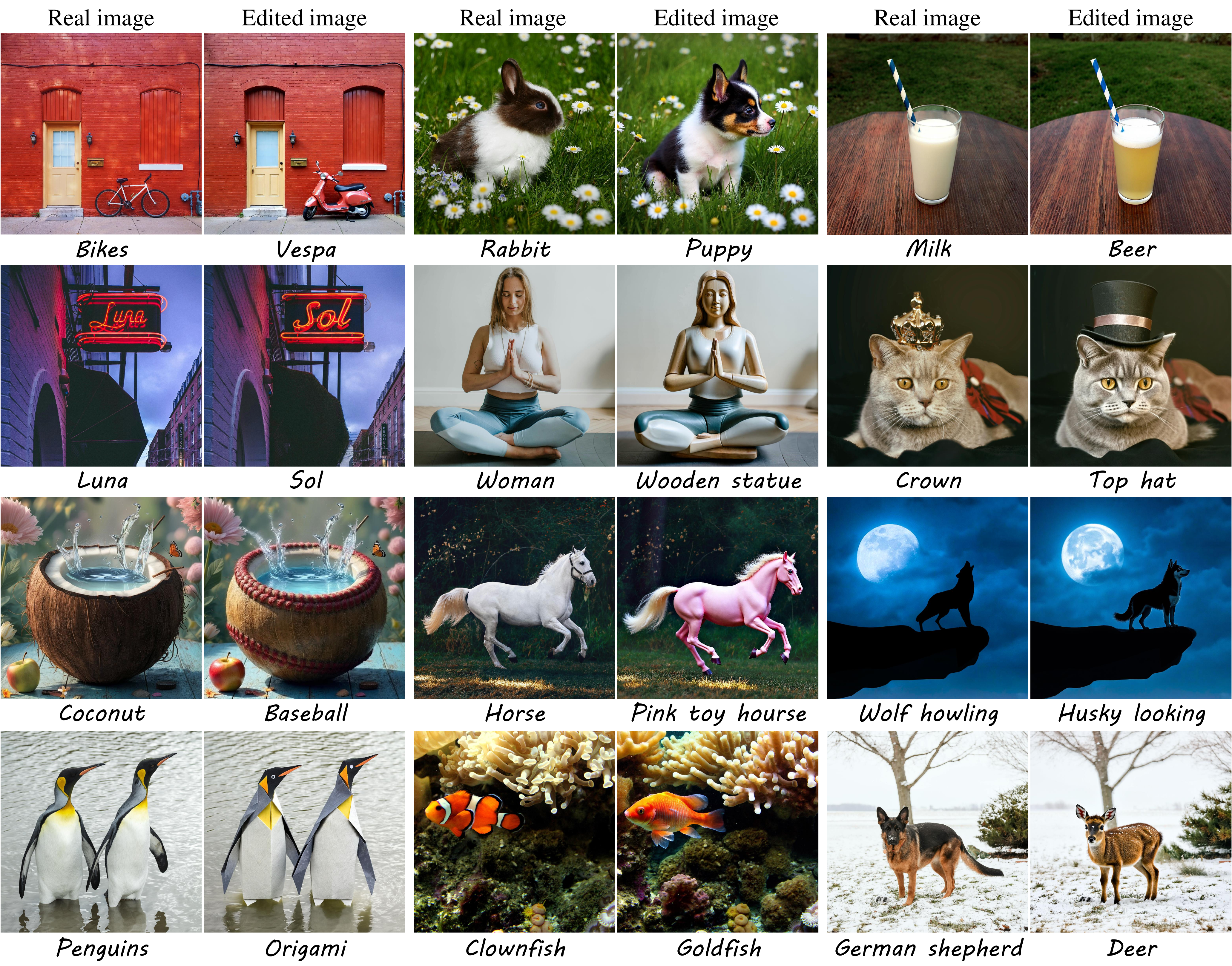}
    \caption{\textbf{\ours{} results.} \ours{} successfully edits diverse images using various editing prompts. The edits preserve the structure of the original image, changing only the specified region. FLUX was used for the first and third rows and SD3 for the second and fourth rows.}
    \label{fig:results}
\end{figure*}

\begin{figure*}
    \centering
    \includegraphics[width=\textwidth]{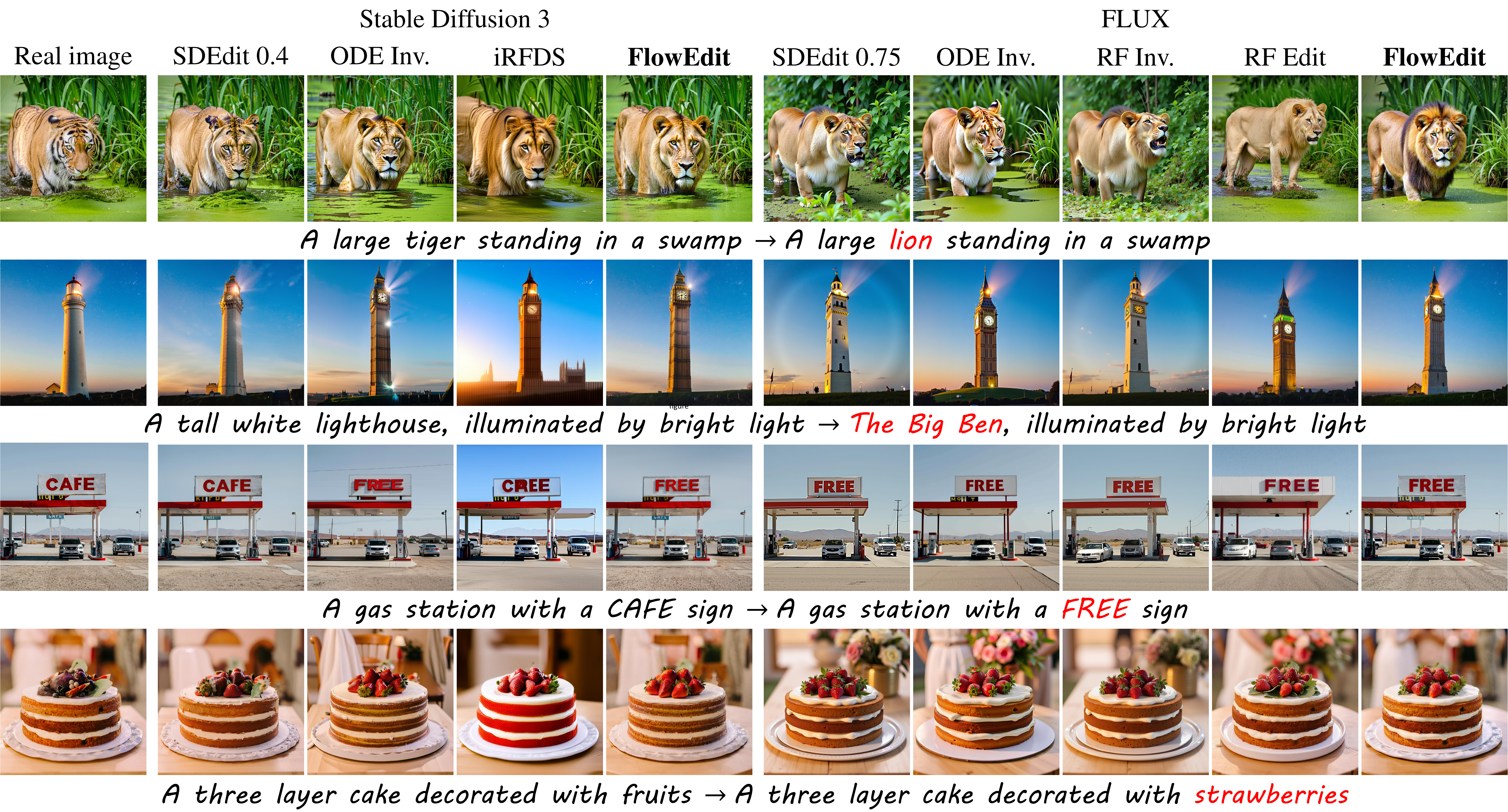}
    \caption{\textbf{Qualitative comparisons.} With both SD3 (left) and FLUX (right), \ours{} better adheres to the target prompt while simultaneously preserving the structure of the original image. The value next to SDEdit indicates the strength.}
    \label{fig:comparisons}
\end{figure*}

\subsection{Relation to optimization based methods}
\label{subsec:relation_optim}
Superficially, \ours{} may seem similar to optimization based methods like DDS \cite{hertz2023delta} and PDS \cite{koo2024posterior}. These techniques, introduced for diffusion rather than flow, attempt to minimize a loss that measures the alignment between the model's noise predictions with the source and target prompts. They approximate the gradient of the loss using only forward-passes through the model. For example, in DDS, the update step is the difference between the two noise predictions, which can be considered similar to our $V^\Delta_t(\hat{Z}^{\text{src}}_{t},\hat{Z}^{\text{tar}}_t)$ direction. However, as we show in App.~\ref{sm:dds}, writing $V^\Delta_t(\hat{Z}^{\text{src}}_{t},\hat{Z}^{\text{tar}}_t)$ in terms of noise predictions reveals that it is different from both the DDS and the PDS updates.

One may still wonder whether \ours{} can be viewed as an optimization process, only for a different loss. Specifically, similarly to DDS, which attempts to minimize the norm of the difference between noise predictions in diffusion models, maybe \ours{} attempts to minimize the norm of the difference between velocity predictions in flow models? In App.~\ref{sm:dds}, we show that this optimization viewpoint is unjustified even for DDS. 
Namely, the DDS iterations do not decrease the DDS loss; they rather tend to \emph{increase} it. Moreover, if allowed to continue beyond the default number of iterations, the quality of the edited image deteriorates. The optimization viewpoint is unnatural also because unlike standard optimizers, \ours{} does not choose timesteps at random but rather uses a monotonically decreasing schedule $\{t_i\}^{n_{\max}}_{i=0}$. Additionally, it must use a learning rate of exactly $dt=t_{i-1}-t_{i}$ at iteration $i$ (namely, the sum of all step-sizes must be 1). Any slight deviation from this learning rate causes the results to deteriorate significantly (see App.~\ref{sm:dds}).

\subsection{Comparison to editing by (exact) inversion}
\label{sec:cats_dogs_exp}
To demonstrate the reduced transport cost of \ours{} compared to editing-by-inversion, we evaluate both methods on a synthetic dataset of model-generated images. 
This way, the initial noise maps are known, ensuring the inversion is exact and eliminating potential issues of approximate inversion. The dataset consists of 1000 cat images generated by SD3 using variations of the prompt ``a photo of cat'' generated by Llama3~\citep{dubey2024llama3}.
We edit these images using both FlowEdit and inversion, with the target prompt identical to the source prompt, except for replacing ``cat'' with ``dog''. We calculate the transport cost for both methods by measuring the MSE between source and edited images in the model's latent space, as well as LPIPS~\cite{zhang2018unreasonable} on the decoded images. 
As expected, \ours{} achieves a lower transport cost (1376 vs.~2239 for MSE, 0.15 vs.~0.25 for LPIPS), indicating superior preservation of the structure and semantics of the source image. 
Figure~\ref{fig:cats_dogs} shows a small qualitative comparison for the cat-to-dog edits.

To assess the alignment of our edits with the target distribution, we generated 1000 dog images with SD3, using the same target prompts but with ``dog'' instead of ``cat''. We followed by calculating FID~\citep{heusel2017fid} and KID~\citep{bińkowski2018kid} between the generated dog images and the edited dog images, for both methods. Our method achieves lower FID (51.14 vs.~55.88) and KID (0.017 vs.~0.023), indicating that similarly to inversion, our ODE path is able to produce images from the target distribution.
See App.~\ref{sm:cats_dogs} for more details and results.

%% file: sec/5_evaluation.tex
\section{Experiments}
\label{sec:evalution}
\paragraph{Implementation details.}

\begin{figure}
    \centering
    \includegraphics[width=0.9\columnwidth]{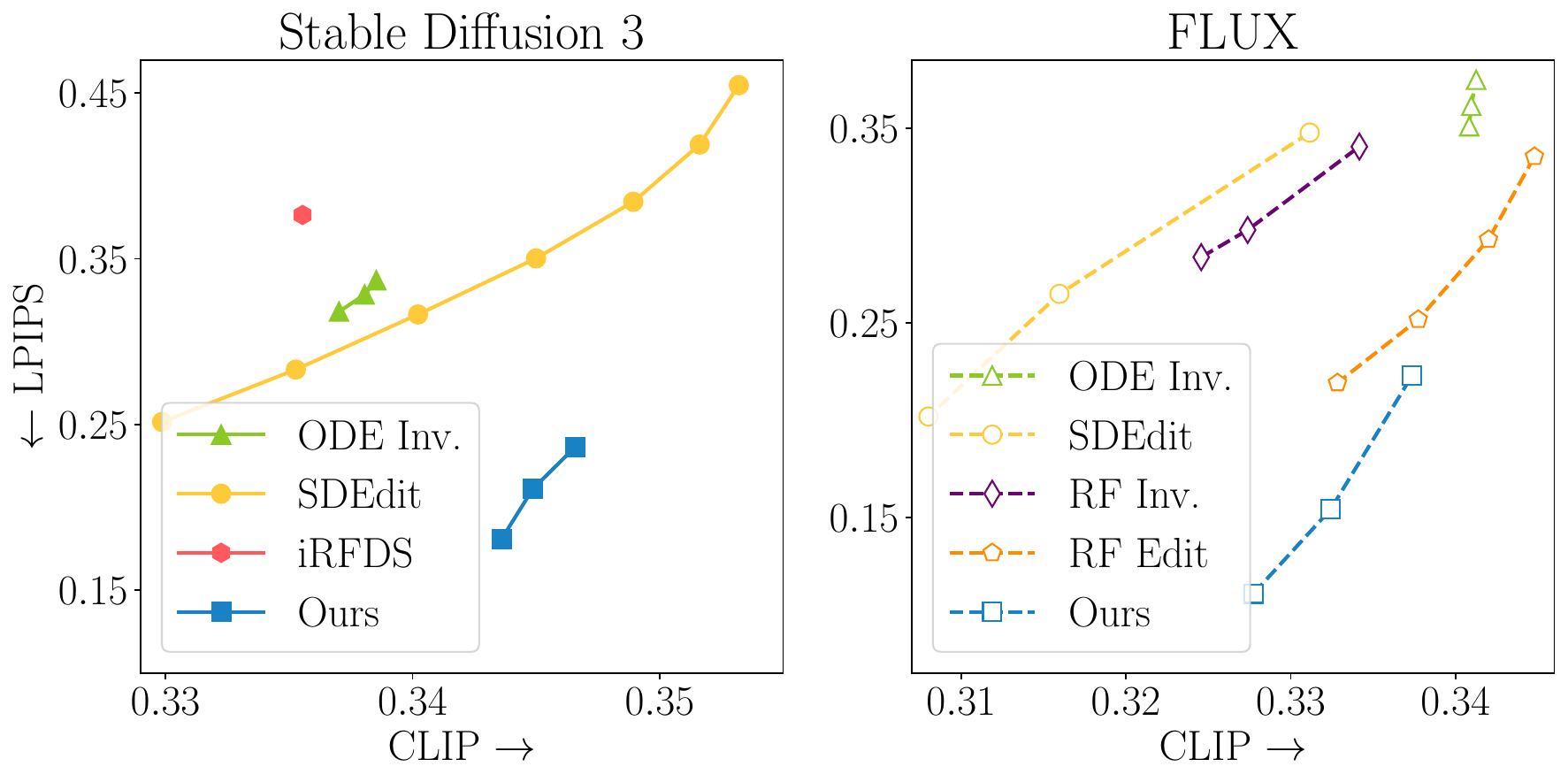}
    \caption{\textbf{Quantitative comparisons.} \ours{} achieves a favorable balance between text adherence (CLIP) and structure preservation (LPIPS) compared to other methods. Connected markers represent different hyperparameters (see App.~\ref{sm:comp}).}
    \label{fig:metrics}
\end{figure}

In our experiments we use the official weights of SD3 medium~\cite{sd3-medium-weights} and FLUX.1 dev~\cite{flux.1-dev-weights}, available at HuggingFace, as the base T2I flow models. 
For SD3 we use $T=50$ steps, with $n_{\text{max}}=33$. SD3 employs CFG \cite{ho2021classifier} for their text conditioning. We set the source and target scales to $3.5$ and $13.5$, respectively. 
For FLUX we use $T=28$ steps, with $n_{\text{max}}=24$. FLUX takes CFG as an input conditioning, which we set to the values of $1.5$ for the source conditioning, and $5.5$ for the target. 
For both methods we use $n_{\text{avg}}=1$ (see Sec.~\ref{sec:pratical} and App.~\ref{sm:practical_navg}). 
We use these hyperparameters for all the results in Figs.~\ref{fig:results},\ref{fig:comparisons},\ref{fig:metrics}. 

\paragraph{Dataset.}

Our dataset consists of a diverse set of over 70 real images of dimension $1024^2$ from the DIV2K dataset~\citep{agustsson2017ntire} and from royalty free online sources~\cite{pexels, pxhere}. Each image has a source prompt, which was obtained from LLaVA-1.5~\cite{liu2024llava15} and manually refined. The source prompts have little effect on \ours{}'s results. In particular, they can be omitted (App.~\ref{sm:source_prompt}). For each image, several handcrafted target prompts are provided to facilitate diverse edits. Overall, the dataset consists of over 250 text-image pairs, and is used to evaluate our and the competing methods. The dataset, including source and target prompt pairs, is available on our official \href{https://github.com/fallenshock/FlowEdit}{github repository}.

\paragraph{Competing methods.}
We compare our method against several competing text-based real image editing methods that use flow models. The first is editing by ODE inversion, which we apply with the same hyperparameters as our method.
The second is SDEdit~\citep{meng2021sdedit}, which is easily applied to flow models by adding noise to the source image up to a specified $n_{\text{max}}$ step and then performing regular sampling conditioned on a target prompt to obtain the edit. The parameter $n_{\text{max}}$ controls the strength of the edit. 
Additionally, we compare to iRFDS \citep{yang2024text}, which is an SDS-based editing method for flow models. We use their official implementation and hyperparameters, which are available only for SD3.     
Furthermore, we compare to RF-Inversion~\citep{rout2024rfinversion}. As an official implementation is not available at the time of writing, we implement it based on the provided pseudo-code and the hyperparameters reported in the SM, which are available only for FLUX. 
Finally, we compare to RF Edit~\cite{wang2024taming} using their official implementation, which is also available only for FLUX. 
Whenever possible, we performed a hyperparameters search for each method to identify the optimal settings for editing. The final parameters used to create the edited images in the paper are detailed in App.~\ref{sm:comp}. 
We do not compare \ours{} to text-based image editing techniques for diffusion models~\cite{huberman2024edit, tumanyan2023plug, cao2023masactrl}, as these are not easily adapted to flow models, and a direct comparison would not be fair due to model differences. 

\paragraph{Qualitative evaluation.}
Figures \ref{fig:teaser},~\ref{fig:results},~\ref{fig:results_sm} show editing results obtained with \ours{} across a diverse set of images and target prompts, including localized edits to text and objects, modifications of multiple objects at once, pose changes, etc. Our edits exhibit good structural preservation of the source image, and simultaneously maintain good adherence to the target text. Figure \ref{fig:comparisons} shows comparisons between \ours{} and the competing methods for both SD3 and FLUX. \ours{} is the only method that consistently adheres to the text prompt while preserving the structure of the original image. For example, \ours{} is the only method that can both change the large text on the gas station sign (third row) and preserve the cars and background of the source image. Some methods, like RF-Inversion, are able to change the text on the sign but fail to maintain the structure of the original image, while others, such as iRFDS, somewhat preserve the structure but fail to fully change the sign. 
In the last row, while all methods perform the required edit (changing the decoration to strawberries), only \ours{} (for both SD3 and FLUX) preserves the background accurately. Other methods introduce additional, unintended elements such as furniture and flowers. 
See App.~\ref{sm:comp} for more qualitative comparisons. 

\paragraph{Quantitative evaluation.} 
We numerically evaluate the results of \ours{} and the other methods using LPIPS~\cite{zhang2018unreasonable} to measure the semantic structure preservation (lower is better) and CLIP~\cite{radford2021clip} to assess text adherence (higher is better). 
Figure \ref{fig:metrics} displays these results for SD3 (left) and FLUX (right), with varied hyperparameters. The plots show that \ours{} achieves a favorable balance between structure preservation and text adherence. The other methods either maintain the structure of original image at the cost of a weak edit or modify the image with little regard to its original semantics.
Additional quantitative evaluations and details, including additional similarity metrics beyond LPIPS, are provided in App.~\ref{sm:comp}.

\paragraph{Text-based style editing.}

\begin{figure}
    \centering
    \includegraphics[width=0.95\linewidth]{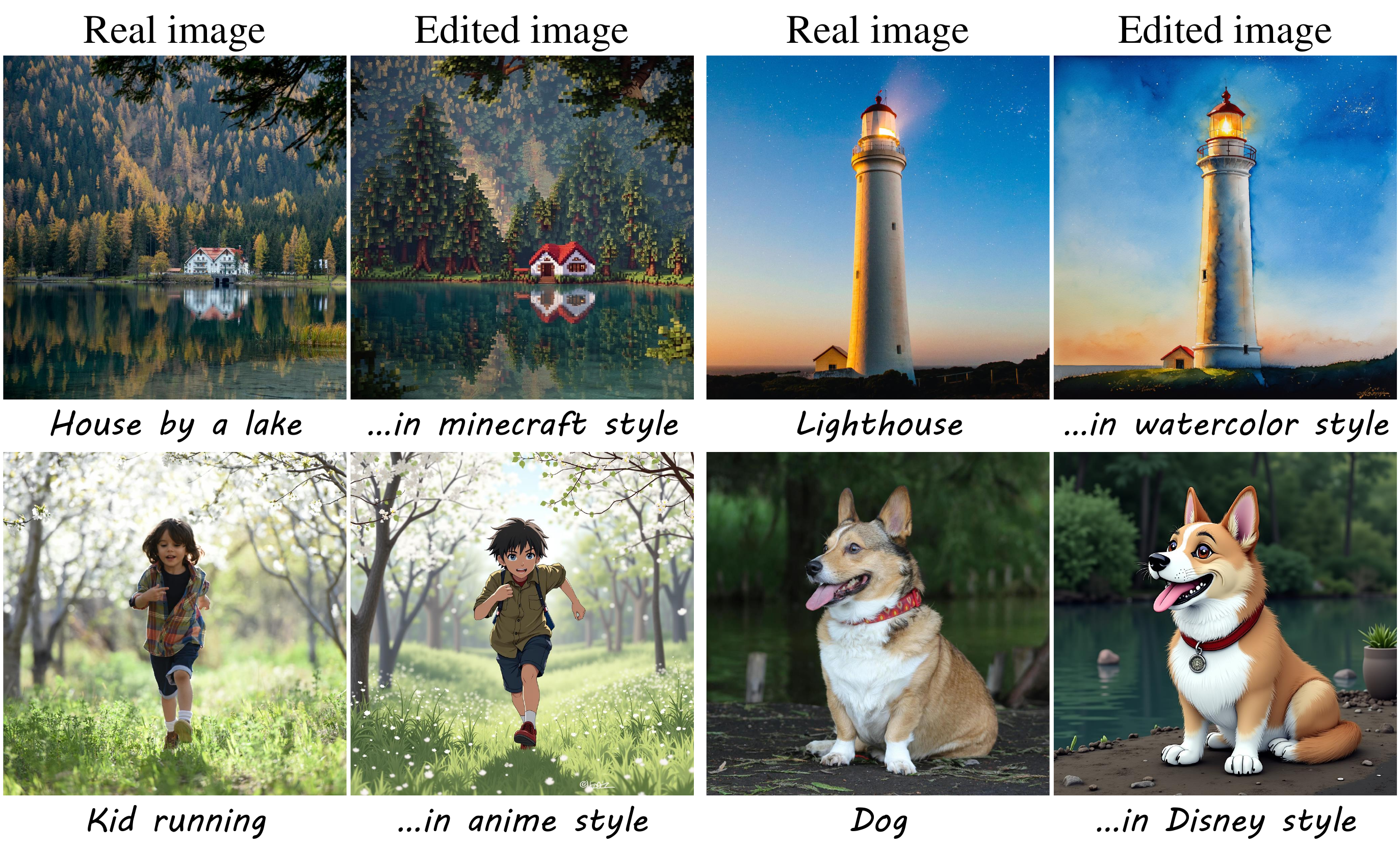}
    \caption{\textbf{Text-based style editing.} \ours{} changes the style of an image at the cost of slightly deviating from the original image structure. SD3 was used for the first row and FLUX for the second. Hyperparameters used for these results are reported in App.~\ref{sm:style}.}
    \label{fig:style}
\end{figure}

The hyperparameters used for the previous experiments lead to strong structure preservation and are thus not optimal for style editing. Figure~\ref{fig:style} demonstrates the results of \ours{} for text-based style editing, where by allowing some deviation from the original structure, we gain stylistic flexibility. 
To achieve these results, we simply remove the dependence on the source image in the final generation steps, as discussed in App.~\ref{sm:style}.

%% file: sec/6_conclusion.tex
\section{Conclusion and limitations}
\label{conclusion}

We introduced \ours{} -- an inversion-free, optimization-free and model agnostic method for text-based image editing using pre-trained flow models. Our approach constructs a direct ODE between the source and target distributions (corresponding to source and target text prompts), without passing through the standard Gaussian distribution as in inversion-based editing. Evaluations on synthetic datasets show that \ours{} achieves lower transport costs, and thus stronger structure preservation. This translates to state-of-the-art performance across various editing tasks, as we illustrated with FLUX and SD3. While \ours{}'s strong structure preservation is beneficial for precise editing tasks, it can become a limitation when substantial modifications to large regions of the image are desired. This is illustrated in Fig.~\ref{fig:limitations} in the context of pose and background editing.

%% file: sec/X_suppl.tex
\section{Additional results}
\label{sm:res}
\begin{figure}[h]
    \centering
    \includegraphics[width=0.97\textwidth]{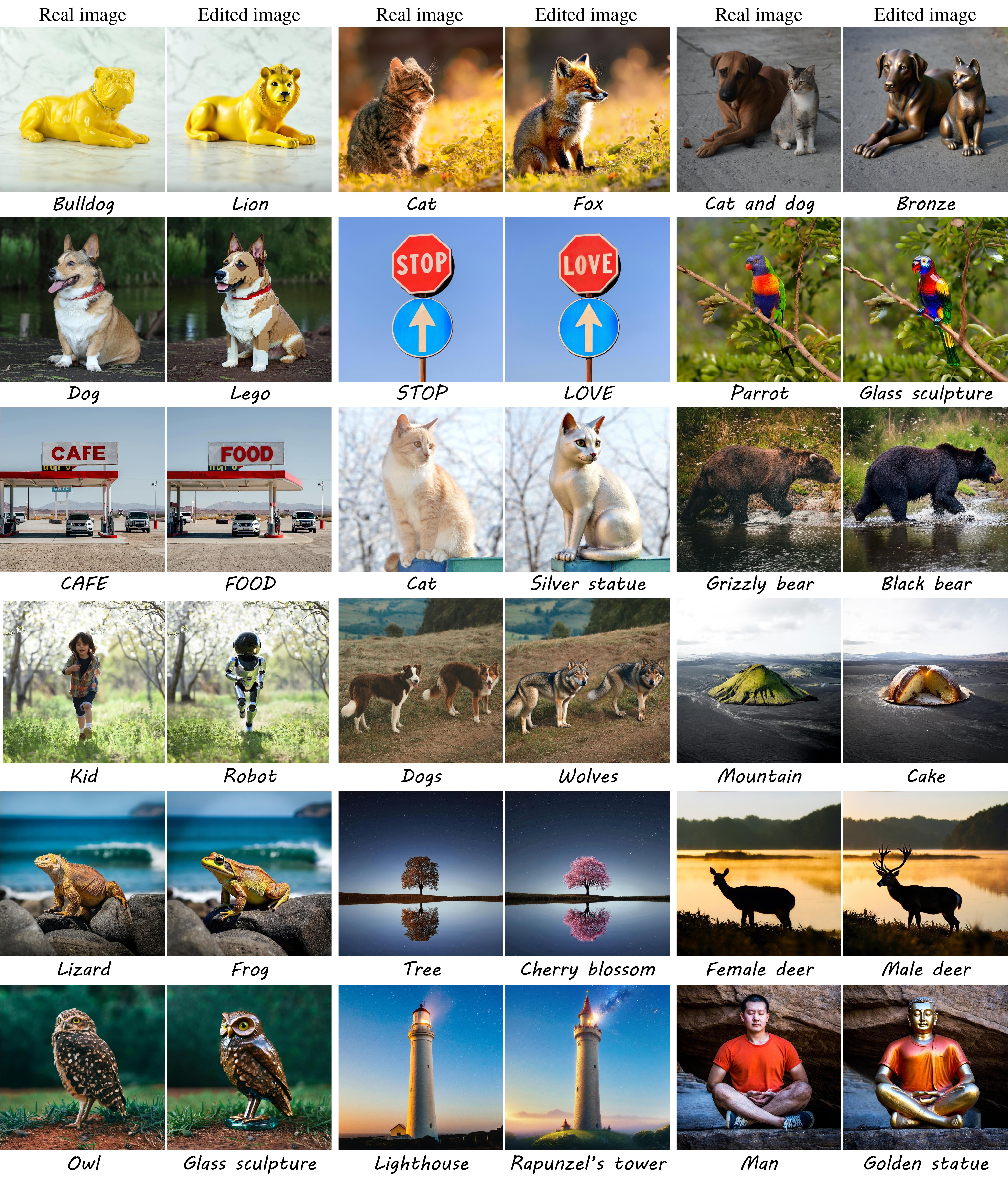}
    \caption{\textbf{Additional \ours{} results.} FLUX was used for the first, third, and fifth rows, and SD3 for the second, fourth and sixth rows.}
    \label{fig:results_sm}
\end{figure}

Figure \ref{fig:results_sm} shows additional editing results obtained with \ours{}.

\clearpage
\section{Comparisons}
\label{sm:comp}
\subsection{Additional qualitative comparisons}

\begin{figure}[h]
    \centering
    \includegraphics[width=\textwidth]{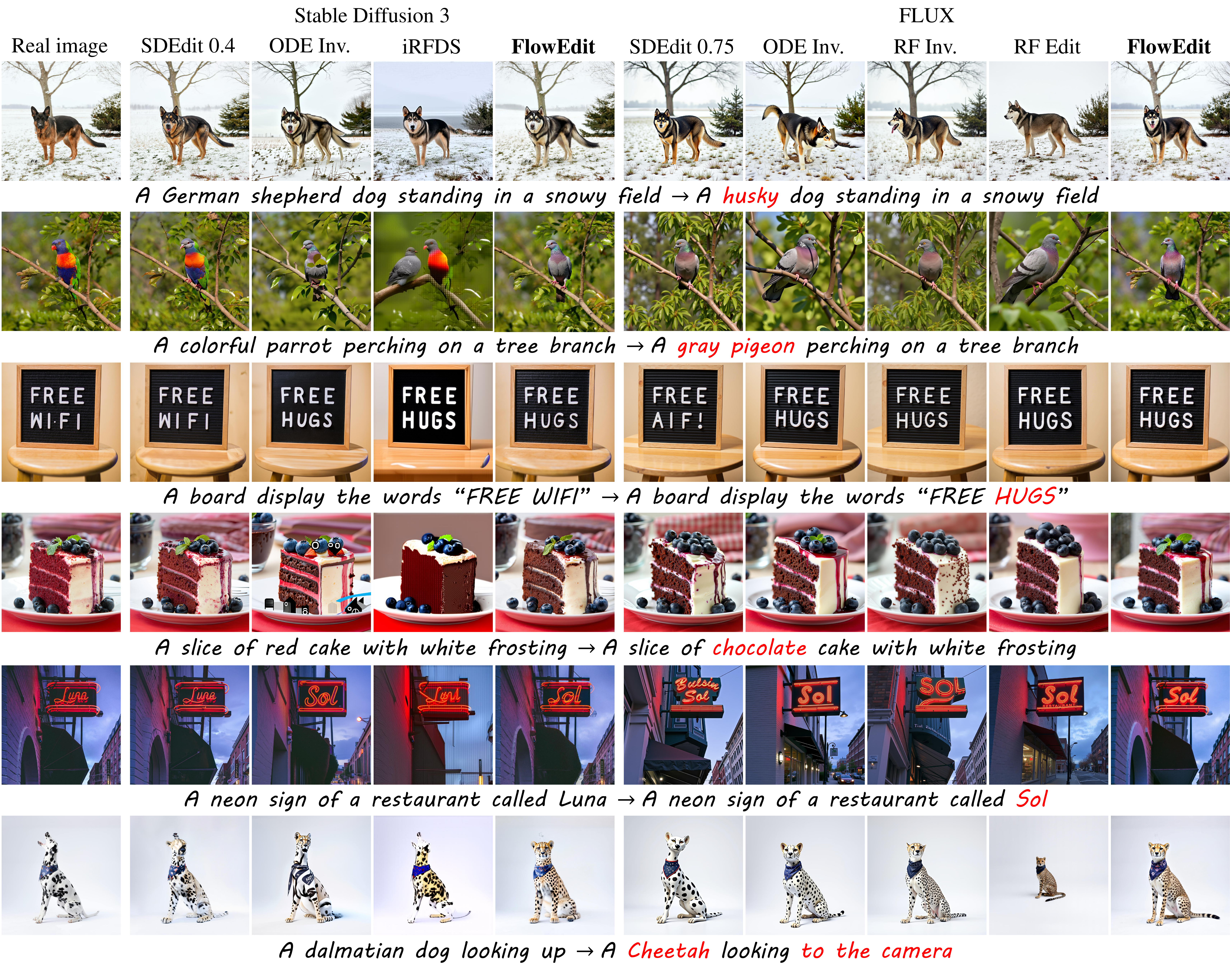}
    \caption{\textbf{Additional qualitative comparisons.}}
    \label{fig:comparisons_sm}
\end{figure}

Figure \ref{fig:comparisons_sm} shows additional comparisons between \ours{} and the competing methods with both SD3 (left) and FLUX (right). The value next to SDEdit indicates the editing strength (see App.~\ref{app:exp_settings}).

\clearpage
\subsection{Additional details on the experiment settings}
\label{app:exp_settings}

We compare \ours{} against the baseline methods of ODE inversion (Sec.~\ref{sec:ode_inversion}) and SDEdit using both FLUX and SD3, as well as against RF-Inversion, RF Edit (FLUX) and iRFDS (SD3). We were not able to compare to Stable Flow~\cite{avrahami2024stable}, as their method is not compatible with Nvidia RTX A6000 (which we used for running \ours{}, as well as all the other methods) even with cpu offloading and when changing the image size from $1024 \times 1024$ to $512 \times 512$.  

Figure \ref{fig:metrics} in the main text presents the CLIP vs.~LPIPS results for \ours{} and the competing methods with both SD3 and FLUX. Below, we detail the hyperparameters used for SD3 (Tab.~\ref{tab:hyperparams-sd3}) followed by those for FLUX (Tabs.~\ref{tab:hyperparams-flux},~\ref{tab:hyperparams-rf-inv},~\ref{tab:hyperparams-rf-edit}). 
The results shown in Figs.~\ref{fig:comparisons}, ~\ref{fig:comparisons_sm} and Tabs.~\ref{tab:metrics-sd3} and~\ref{tab:metrics-flux} were obtained using the parameters listed in Tabs.~\ref{tab:hyperparams-sd3},~\ref{tab:hyperparams-flux} and ~\ref{tab:hyperparams-rf-inv}. The bold entries indicate the specific settings used when multiple hyperparameter options were tested.

\subsubsection{Stable Diffusion 3 hyperparameters}

In Fig.~\ref{fig:metrics} in the main text both \ours{} and ODE Inversion are shown with three options for the CFG target scale, as detailed in Tab.~\ref{tab:hyperparams-sd3}, from left to right. For SDEdit the different values of $n_{\text{max}}$ represents different strength settings ranging from $0.2$ to $0.8$ in intervals of $0.1$. The markers in the figure indicate results with these strength values from left to right. 

\begin{table*}[h]
\centering
\caption{\textbf{SD3 hyperparameters.}}
\label{tab:hyperparams-sd3}
\begin{tabular}{@{}ccccc@{}}
\toprule
 & $T$ steps & $n_{\text{max}} $ & CFG @ source & CFG @ target \\ \midrule
SDEdit & $50$ &  $10,\;15,\;\textbf{20},\;25,\;30,\;35,\;40$ & -  & 13.5  \\
ODE Inv. & $50$ & $33$ & $3.5$ & $\textbf{13.5},\;16.5,\;,19.5$     \\
iRFDS & \multicolumn{4}{c}{official implementation and hyperparameters} \\
\midrule
\ours{} & $50$ & $33$ & $3.5$ & $\textbf{13.5},\;16.5,\;,19.5$     \\
\bottomrule
\end{tabular}
\end{table*}

\subsubsection{FLUX hyperparameters}

In Fig.~\ref{fig:metrics} in the main text both \ours{} and ODE Inversion are shown with three options for the CFG target scale, as detailed in Tab.~\ref{tab:hyperparams-flux}, from left to right. 

For SDEdit the different values of $n_{\text{max}}$ represent different strength values, corresponding to $0.25,\;0.5,\;0.75$. The markers in the figure indicate results with these strength values from left to right. 

\begin{table*}[h]
\centering
\caption{\textbf{FLUX hyperparameters.}}
\label{tab:hyperparams-flux}
\begin{tabular}{@{}ccccc@{}}
\toprule
 & $T$ steps & $n_{\text{max}} $ & CFG @ source & CFG @ target \\ \midrule
SDEdit & $28$ &  $7,\;14,\;\textbf{21}$ & -  & $5.5$  \\
ODE Inv. & $28$ & $20,\;\textbf{24}$ & $1.5$ & $3.5,\;4.5,\;\textbf{5.5}$     \\
\ours{} & $28$ & $24$ & $1.5$ & $3.5,\;4.5,\;\textbf{5.5}$     \\
\bottomrule
\end{tabular}
\end{table*}

For RF-Inversion, we explore multiple sets of hyperparmeters, as the paper does not report specific ones for general editing. Following the SM of their work, we experimented with several combinations, detailed in Tab.~\ref{tab:hyperparams-rf-inv} using their notations.

\begin{table*}[h]
\centering
\caption{\textbf{RF-Inv. hyperparameters.}}
\label{tab:hyperparams-rf-inv}
\begin{tabular}{@{}ccccc@{}}
\toprule
$T$ steps & $s$ starting time & $\tau$ stopping time & $\eta$ strength \\ \midrule
$28$ &  $0$ & $8,\;7,\;\textbf{6}$  & $\textbf{0.9}, 1.0$  \\
\bottomrule
\end{tabular}
\end{table*}

For RF Edit, we explore multiple injection scales as the paper does not report the injection scale used for image editing. All hyperparameters used for RF Edit are listed in Tab.~\ref{tab:hyperparams-rf-edit} using their notations.

\begin{table*}[h]
\centering
\caption{\textbf{RF Edit. hyperparameters.}}
\label{tab:hyperparams-rf-edit}
\begin{tabular}{@{}cccc@{}}
\toprule
Steps & Guidance & Injection \\ \midrule
$30$ &  $2$ & $\textbf{2},\;3,\;4,\;5$   \\
\bottomrule
\end{tabular}
\end{table*}

\begin{figure}[h]
    \centering
    \includegraphics[width=0.5\textwidth]{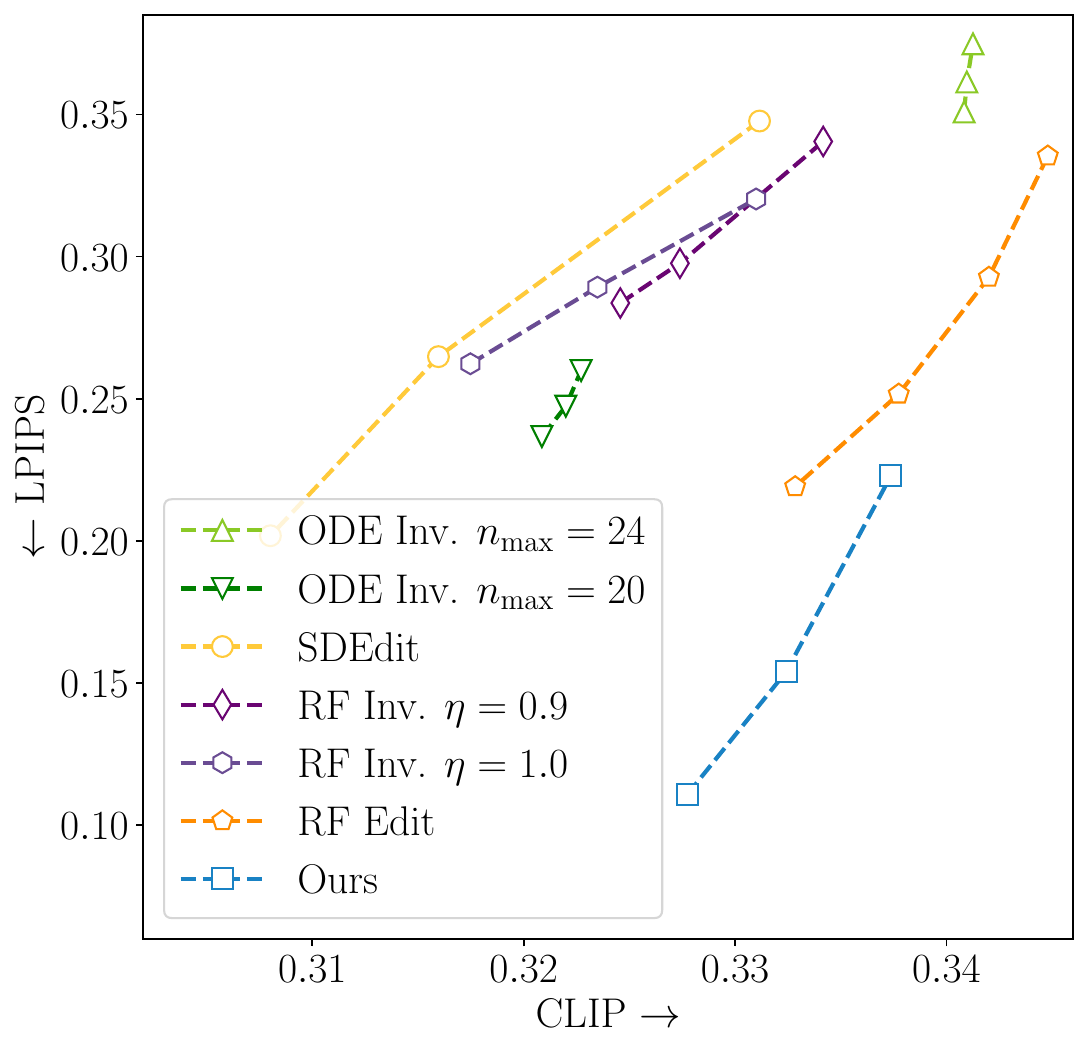}
    \caption{\textbf{Additional quantitative comparisons.} In addition to the results shown in Fig.~\ref{fig:metrics} for FLUX, we include additional hyperparameters: ODE Inversion with $n_{\text{max}}=20$ and RF-Inversion with $\eta=1.0$. These configurations also struggle to achieve a good balance between the CLIP and LPIPS metrics. In contrast, \ours{} demonstrates the best balance.}
    \label{fig:metrics_flux_sm}
\end{figure}

Lastly, as can be seen in Fig.~\ref{fig:metrics}, ODE Inversion on FLUX with these hyperparameters achieves a high CLIP score at the cost of a high LPIPS score, indicating it does not balance these metrics effectively. By varying $n_{\text{max}}$, ODE Inversion could achieve lower (better) LPIPS scores at the cost of lower (worse) CLIP scores. Figure \ref{fig:metrics_flux_sm} illustrates the CLIP and LPIPS scores for the different methods using FLUX. 
Specifically, ODE Inversion is also shown with $n_{\text{max}}=20$ in addition to $n_{\text{max}}=24$ as shown in the main text. However, with these adjusted hyperparameters, ODE Inversion struggles to adhere to text prompts. 
RF-Inversion results with $\eta=1.0$ are also illustrated in Fig.~\ref{fig:metrics_flux_sm} and Tab.~\ref{tab:metrics-flux}. Again, with this hyperparameter this method struggles to adhere to text prompts. 

\clearpage
\subsection{Metrics comparisons}
\label{app:metrics_comparisons}

In addition to quantifying the structure preservation using LPIPS, as described in the main text, we now report alternative metrics. 
Specifically, we use DreamSim~\cite{fu2024dreamsim} as well as cosine similarity in the CLIP~\cite{radford2021clip} and DINO~\cite{caron2021dino} embedding spaces, between the source and target images' embeddings. 
For DreamSim, a lower score indicates better structure preservation. For CLIP images and DINO, a higher score (bounded by 1) means higher structure preservation. 

Tables~\ref{tab:metrics-sd3},~\ref{tab:metrics-flux} illustrate the results of these metrics, alongside LPIPS and CLIP for the hyperparameters described above.  As can be seen, \ours{} is the only method that is able to both adhere to the text prompt and preserve the structure of the source image. 

\begin{table*}[h]
\centering
\caption{\textbf{SD3 metrics.} The \colorbox{tabfirst}{first}, \colorbox{tabsecond}{second} and \colorbox{tabthird}{third} best scores are highlighted for each metric. CLIP-T measures adherence to text, while the other four scores measure structure preservation.}
\label{tab:metrics-sd3}
\begin{tabular}{@{}c|c|cccc@{}}
\toprule
 & CLIP-T $\uparrow$ & CLIP-I $\uparrow$ & LPIPS $\downarrow$ & DINO $\uparrow$ & DreamSim $\downarrow$ \\ \midrule
SDEdit 0.2 & 0.33  & \colorbox{tabfirst}{0.885}  & \colorbox{tabsecond}{0.251}  & \colorbox{tabsecond}{0.634} & \colorbox{tabfirst}{0.213}  \\
SDEdit 0.4 & \colorbox{tabsecond}{0.34}  & \colorbox{tabthird}{0.854}  & \colorbox{tabthird}{0.316}  & \colorbox{tabthird}{0.564} & \colorbox{tabthird}{0.273}  \\
ODE Inv. & \colorbox{tabthird}{0.337}  & 0.813  & 0.318  & 0.549  & 0.326  \\
iRFDS & 0.335 & 0.822 & 0.376 & 0.534 & 0.327
\\
\midrule
\textbf{\ours{}} & \colorbox{tabfirst}{0.344} & \colorbox{tabsecond}{0.872} & \colorbox{tabfirst}{0.181} & \colorbox{tabfirst}{0.719} & \colorbox{tabsecond}{0.253} \\ \bottomrule
\end{tabular}
\end{table*}

\begin{table*}[h]
\centering
\caption{\textbf{FLUX metrics.} The \colorbox{tabfirst}{first}, \colorbox{tabsecond}{second} and \colorbox{tabthird}{third} best scores are highlighted for each metric. CLIP-T measures adherence to text, while the other four scores measure structure preservation.}
\label{tab:metrics-flux}
\begin{tabular}{@{}c|c|cccc@{}}
\toprule
 & CLIP-T $\uparrow$ & CLIP-I $\uparrow$ & LPIPS $\downarrow$ & DINO $\uparrow$ & DreamSim $\downarrow$ \\ \midrule
SDEdit 0.5 & 0.316  &  \colorbox{tabfirst}{0.902} & \colorbox{tabthird}{0.264}  & \colorbox{tabsecond}{0.637}  & \colorbox{tabfirst}{0.18}  \\
SDEdit 0.75 & 0.331  &  0.862 & 0.348  & 0.557  & 0.26 \\
ODE Inv. & \colorbox{tabsecond}{0.341} & 0.822 & 0.374  & 0.505  &  0.328   \\
RF Inv. & 0.334 & 0.856 & 0.34 & 0.558 & 0.266 \\
RF Edit 2 & \colorbox{tabfirst}{0.344} & 0.833 & 0.335 & 0.53 & 0.32 \\
RF Edit 5 & 0.332 & \colorbox{tabsecond}{0.876} & \colorbox{tabfirst}{0.22} & \colorbox{tabthird}{0.65} & \colorbox{tabsecond}{0.22} \\
\midrule
\textbf{\ours{}} & \colorbox{tabthird}{0.337} & \colorbox{tabthird}{0.875} & \colorbox{tabsecond}{0.223} & \colorbox{tabfirst}{0.682} & \colorbox{tabthird}{0.252} \\
\bottomrule
\end{tabular}
\end{table*}

\clearpage
\section{Effect of $n_{\text{max}}$}
\label{sm:n_max}

\begin{figure}[h]
    \centering
    \includegraphics[width=\textwidth]{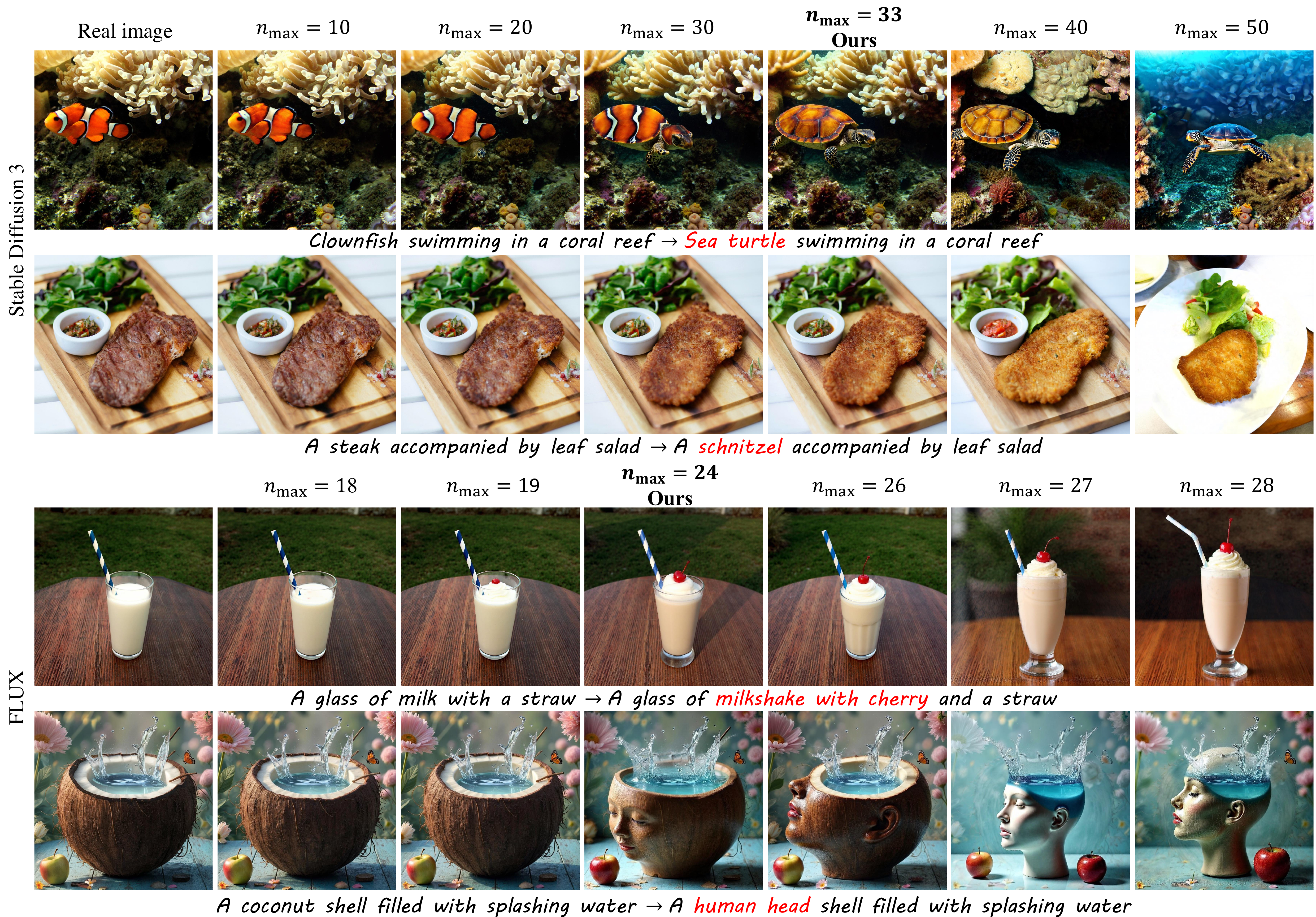}
    \caption{\textbf{Effect of $n_{\text{max}}$.} }
    \label{fig:n_max_effect_sm}
\end{figure}

As described in the main text, we define an integer to determine the starting timestep of the process, where $0 \leq n_{\text{max}} \leq T$. The process is initialized with $\smash{Z^{\text{FE}}_{t_{n_{\text{max}}}}=X^{\text{src}}}$. When $n_{\text{max}}=T$, the full edit path is traversed and the strongest edit is obtained. For $n_{\text{max}}<T$, the first $(T - n_{\text{max}})$ timesteps are skipped, effectively shortening the edit path. This is equivalent to inversion, where weaker edits are obtained by inverting up to timestep $n_{\text{max}}$, and sampling from there. Figure \ref{fig:n_max_effect_sm} illustrates the effect of $n_{\text{max}}$ on the results. The CFG and $T$ used for these editing results are the same as mentioned in the main text, except 
for $n_{\text{max}}$ whose value is specified in the figure. 

\clearpage
\section{Text-based style editing}
\label{sm:style}

\ours{} excels in structure preserving edits. This trait is usually desired in text-based image editing, yet in some cases it could be too limiting. One such scenario is text-based style editing, where we would like to slightly deviate from the original structure in order to achieve a stronger stylistic effect. While $n_{\text{max}}$ allows some control over the structure, it might not be enough by itself to control the finer details required for style changes, \ie higher frequency textures.

To achieve better control over the finer details, we define a new hyperparameter, $n_{\text{min}}$, that controls the structure deviation at lower noise levels, and effectively allows stronger modifications to the higher frequencies. Specifically, when $i < n_{\text{min}}$ we apply regular sampling with the target text, rather than following \ours{} as described in the main text. These steps are further detailed in our full algorithm, Alg.~\ref{alg:full}. 

\begin{figure}[h]
    \centering
    \includegraphics[width=\textwidth]{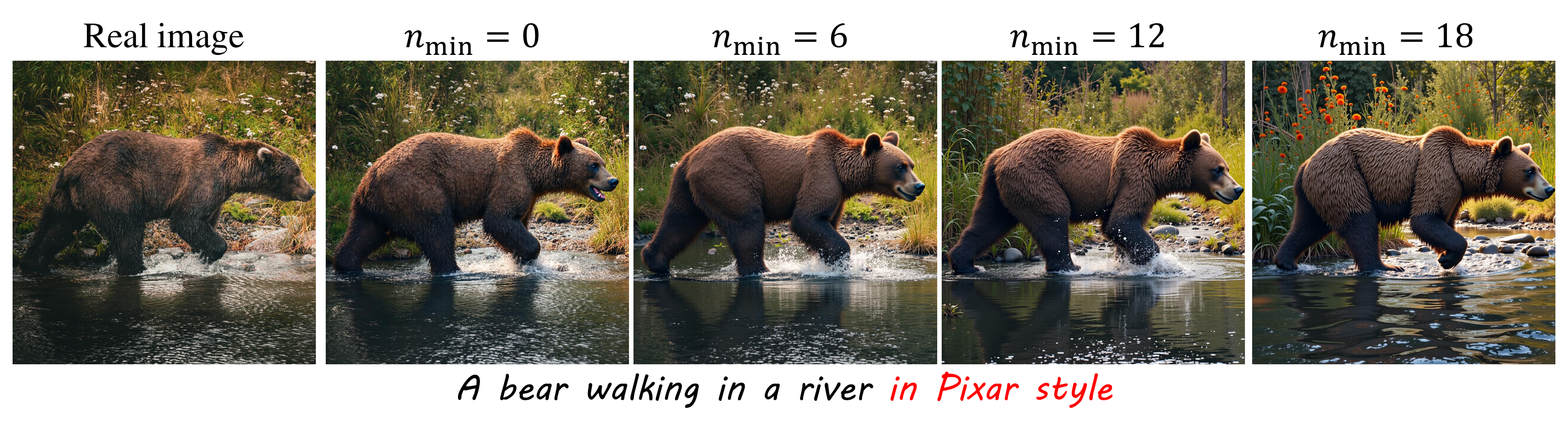}
    \caption{\textbf{Effect of $n_{\text{min}}$.} When $n_{\text{min}}$ is small, the edited image remains close to the original image and struggles to align with the text. In contrast, larger values of $n_{\text{min}}$ result in better adherence to the text but at the cost of greater deviation from the original image. }
    \label{fig:n_min_effect_sm}
\end{figure}

Figure \ref{fig:n_min_effect_sm} illustrates the effect of $n_{\text{min}}$ for text-based style editing. For small $n_{\text{min}}$ the edited image preserves the structure of the original image (especially in the higher frequencies), while for higher $n_{\text{min}}$ values it slightly deviates from it, achieving a stronger edit. These results were obtained using FLUX with $T=28$ steps, $n_{\text{max}}=21$ and CFG scales of $2.5$ and $6.5$ for the source and target conditionings, respectively. The $n_{\text{min}}$ values are mentioned in the figure.

The table below includes the hyperparameters used for the text-based style editing results in Fig.~\ref{fig:style}. 

\begin{table*}[h]
\centering
\caption{\textbf{Hyperparameters used for the results displayed in Fig.~\ref{fig:style}.}}
\label{tab:metrics-style}
\begin{tabular}{@{}cccccc@{}}
\toprule
& Model & $n_{\text{min}}$ & $n_{\text{max}}$ & CFG @ source & CFG @ target \\ \midrule
House by a lake in minecraft style & SD3  & 15 & 31  & 3.5 & 13.5  \\
Lighthouse in watercolor painting style & SD3  & 15 & 31  & 3.5 & 13.5  \\
Kid running in anime style & FLUX  & 14 & 21 & 2.5 & 6.5  \\
Dog in Disney style & FLUX  & 14  & 24 & 1.5 & 4.5  \\
\bottomrule
\end{tabular}
\end{table*}

\clearpage
\section{Hyperparmeters used for Fig.~\ref{fig:teaser}}
\label{sm:hyperparams}

The results in Fig.~\ref{fig:teaser} were achieved using the hyperparmeters described in Tab.~\ref{tab:metrics-fig1} below.

\begin{table*}[h]
\centering
\caption{\textbf{Hyperparameters used for the results displayed in Fig.~\ref{fig:teaser}.}}
\label{tab:metrics-fig1}
\begin{tabular}{@{}cccccc@{}}
\toprule
& Model & $n_{\text{\text{min}}}$ & $n_{\text{\text{max}}}$ & CFG @ source & CFG @ target \\ \midrule
this $\rightarrow$ home & FLUX  & 0 & 24 & 1.5 & 3.5  \\
Cat $\rightarrow$ Raccoon & FLUX  & 0  & 24 & 1.5 & 4.5  \\
LOVE $\rightarrow$ FLOW & FLUX  & 0  & 24 & 1.5 & 4.5  \\
Bread $\rightarrow$ Bacon & FLUX  & 0 & 24 & 1.5 & 3.5  \\
Mountain $\rightarrow$ Volcano & FLUX  & 0 & 24 & 1.5 & 5.5  \\
Lizard $\rightarrow$ Dragon & SD3  & 0  & 33 & 3.5 & 13.5  \\
Man jumping in Pixar style & SD3  & 15 & 21  & 3.5 & 13.5  \\
Bread $\rightarrow$ Steak & SD3  & 0  & 33 & 3.5 & 13.5  \\
White dog w/ cat $\rightarrow$ Dalmatian w/o cat & SD3  & 0  & 33  & 3.5 & 13.5  \\
\bottomrule
\end{tabular}
\end{table*}

\clearpage
\section{Full Algorithm}

\begin{algorithm}[h]
   \caption{Full \ours{} algorithm}
   \label{alg:full}
\begin{algorithmic}
   \STATE {\bfseries Input:} 
   real image $X^{\text{src}},\big\{t_i\big\}^T_{i=0}, n_{\text{max}}, n_{\text{min}}, n_{\text{avg}}$ 
   \STATE {\bfseries Output:} edited image $X^{\text{tar}}$
   \STATE {\bfseries Init:} $Z^{\text{FE}}_{t_{\text{max}}}=X^{\text{src}}_0$
   
   \FOR{$i=n_{\text{max}}$ {\bfseries to } $n_{\text{min}+1}$}
   \STATE $N_{t_i} \sim \mathcal{N}(0,\,1)$\tikzmark{top}
   \STATE $Z^{\text{src}}_{t_i} \leftarrow (1-t_i)X^{\text{src}} + t_iN_{t_i}$
   \STATE $Z^{\text{tar}}_{t_i} \leftarrow Z^{\text{FE}}_{t_i} +Z^{\text{src}}_{t_i}- X^{\text{src}}$
   \STATE $V^{\Delta}_{t_i} \leftarrow V^{\text{tar}}(Z^{\text{tar}}_{t_i},t_i)-V^{\text{src}}(Z^{\text{src}}_{t_i},t_i)$\tikzmark{right}\tikzmark{bottom}
   \STATE $Z^{\text{FE}}_{t_{i-1}} \leftarrow Z^{\text{FE}}_{t_i} + (t_{i-1}-t_{i})V^{\Delta}_{t_i}$
   \ENDFOR
   \IF{$n_\text{min}=0$}
    \STATE {\bfseries Return:} $Z^{\text{FE}}_{0} = X_{0}^{\text{tar}}$
   \ELSE
      \STATE $N_{n_{\text{min}}} \sim \mathcal{N}(0,\,1)$
       \STATE $Z^{\text{src}}_{t_{n_{\text{min}}}} \leftarrow (1-t_{n_{\text{min}}})X^{\text{src}} + t_{n_{\text{min}}}N_{n_{\text{min}}}$
       \STATE $Z^{\text{tar}}_{t_{n_{\text{min}}}} \leftarrow Z^{\text{FE}}_{t_{n_{\text{min}}}} +Z^{\text{src}}_{t_{n_{\text{min}}}}- X^{\text{src}}$

       \FOR{$i=n_{\text{min}}$ {\bfseries to} $1$}
            \STATE $Z^{\text{tar}}_{t_{i-1}} \leftarrow Z^{\text{tar}}_{t_i} + (t_{i-1}-t_{i})V^{\text{tar}}(Z^{\text{tar}}_{t_i}, t_i)$       
       \ENDFOR
       \STATE {\bfseries Return:} $Z^{\text{tar}}_{0} = X_{0}^{\text{tar}}$
   \ENDIF
\end{algorithmic}
\AddNote{top}{bottom}{right}{Optionally average $n_{\text{avg}}$ samples}
\end{algorithm}

\clearpage
\section{Limitations}
\label{sm:limitations}

\begin{figure}[h]
    \centering
    \includegraphics[width=0.7\linewidth]{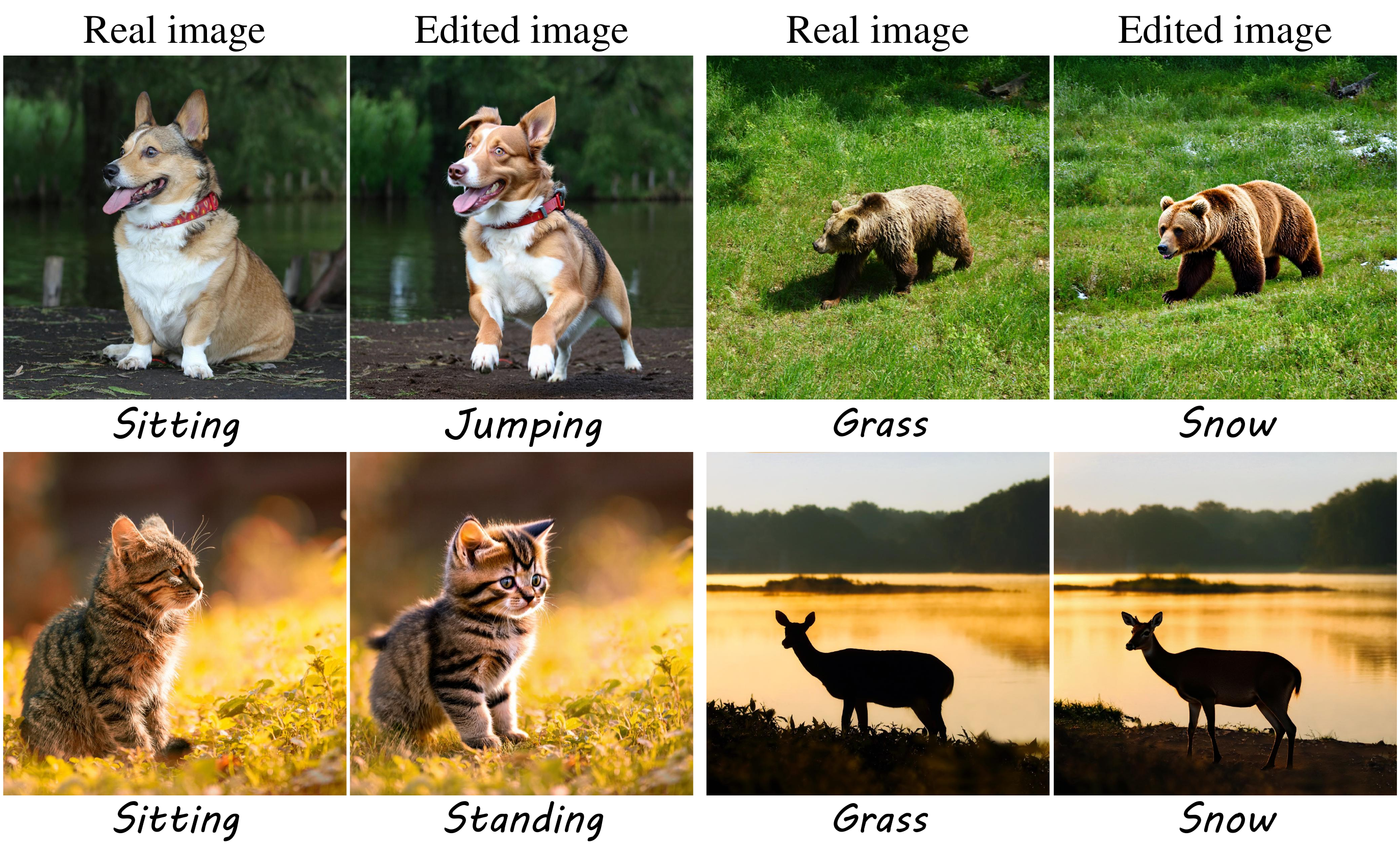}
    \caption{\textbf{Limitations.} \ours{} is often fails at making substantial modifications to large region of the image, as in pose (left) and background (right) editing.}
    \label{fig:limitations}
\end{figure}

\ours{} excels in structure preserving edits, and is therefore often limited in its ability to make substantial modifications to large regions of the image. This is illustrated in Fig.~\ref{fig:limitations} for pose and background editing, respectively. In these cases, \ours{} does not fully modify the image according to the target prompt and it fails at preserving the source identity. 
To obtain a greater deviation from the source image, it is possible to increase $n_{\text{min}}$, as we illustrate in App.~\ref{sm:style}. This is helpful for style editing, but is often still not enough for background and pose edits. 

\clearpage
\section{Effect of random noise on the results}
\label{sm:random}

\begin{figure}[h]
    \centering
    \includegraphics[width=0.7\linewidth]{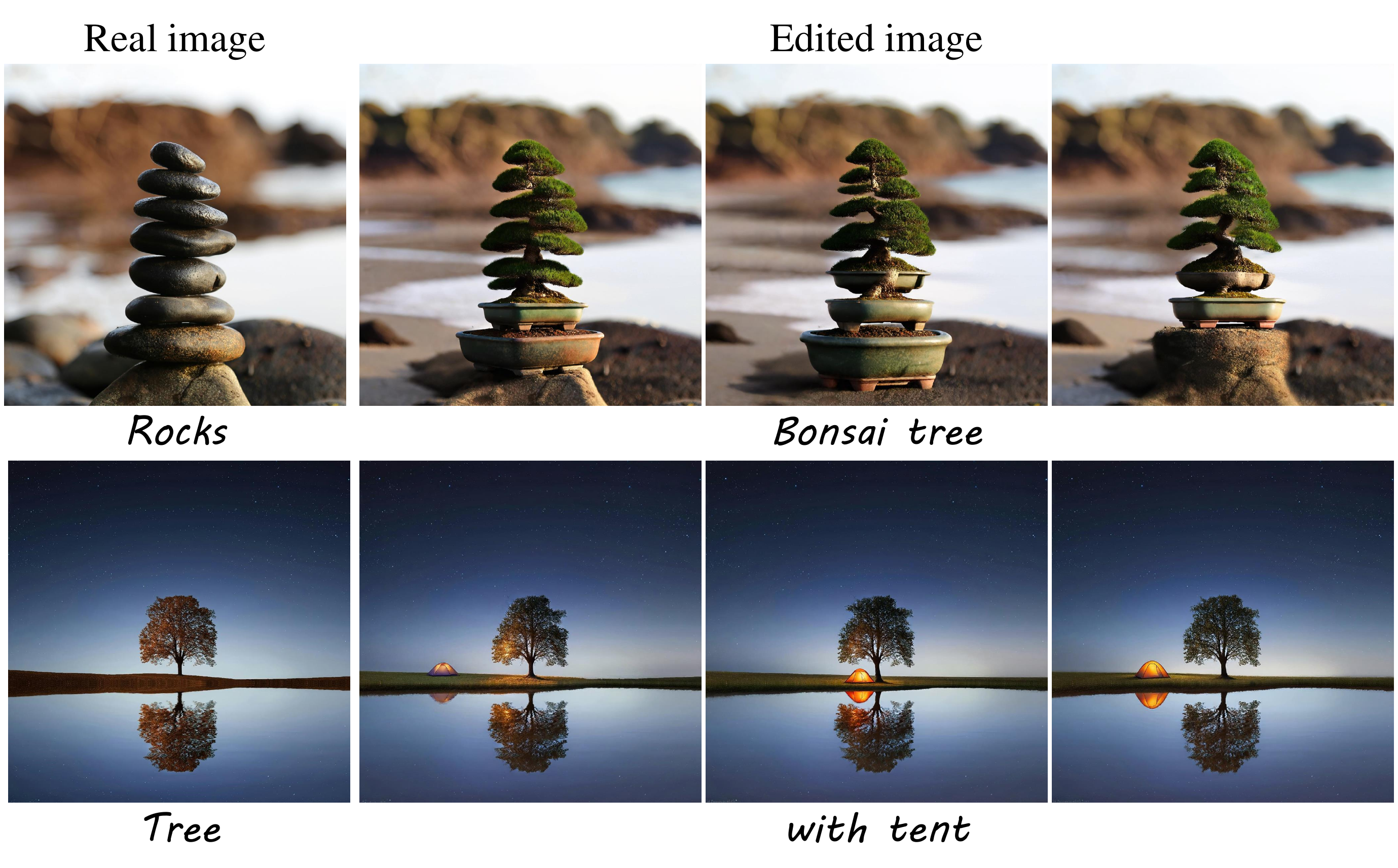}
    \caption{\textbf{\ours{} diverse results due to different added noise.}}
    \label{fig:diversity}
\end{figure}

\ours{} adds random noise to the source image. If the number of samples over which we average is small, then the algorithm is effectively stochastic. Namely, it produces different editing results with different random seeds. These variations can lead to diverse edits for the same text-image pair. As shown in Fig.~\ref{fig:diversity}, the rocks are transformed into different bonsai trees, and the tent appears in various locations within the image.

\begin{figure}[h]
    \centering
    \includegraphics[width=0.7\linewidth]{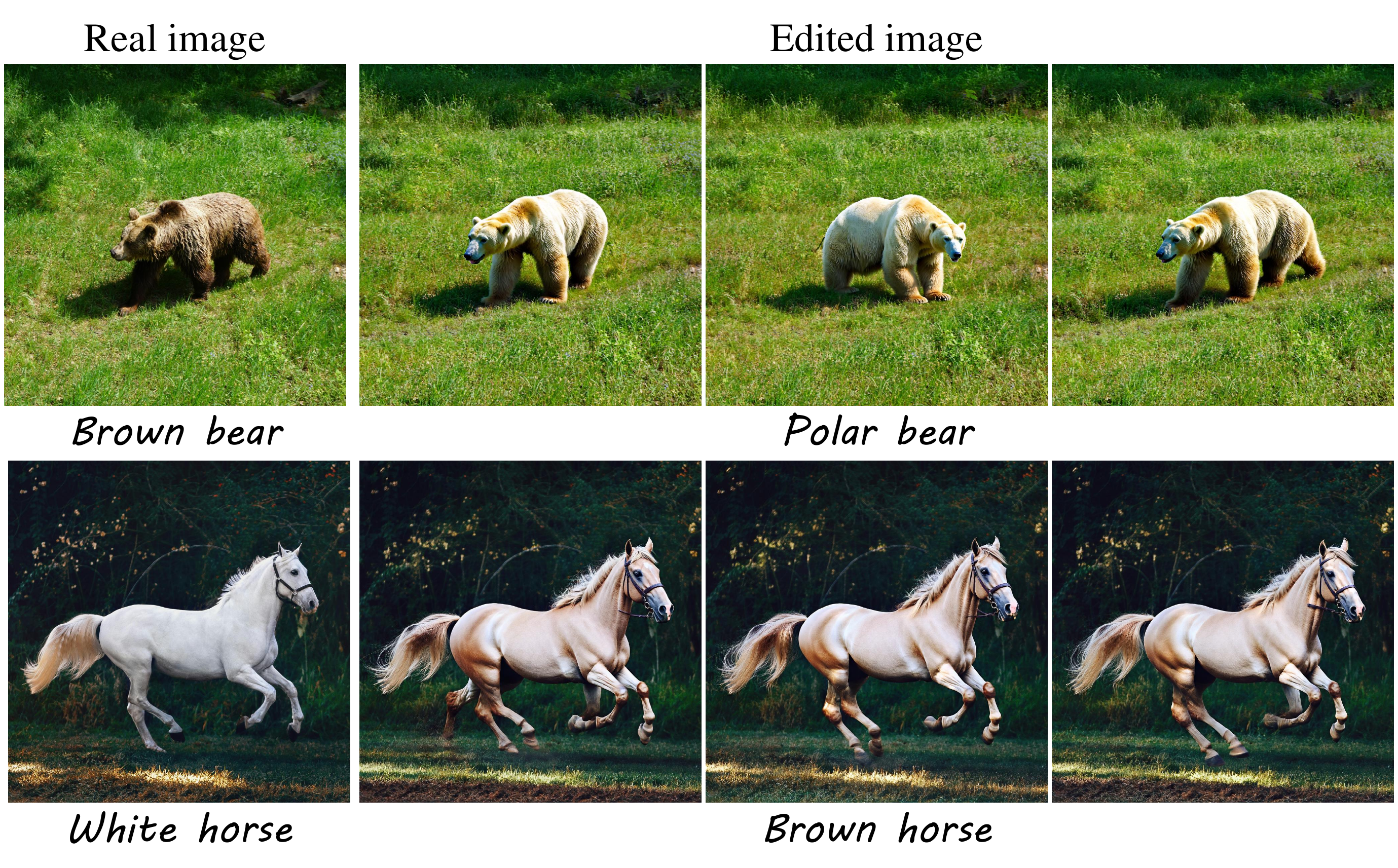}
    \caption{\textbf{Failure cases due to different added noise.}}
    \label{fig:failure}
\end{figure}

However, these changes can also result in failure cases, as illustrated in Fig.~\ref{fig:failure}. For example, when editing a white horse into a brown one, the edited results sometimes show the horse with more than four legs or suffer from other artifacts.

The editing results in both figures were obtained using SD3 and the hyperparameter mentioned in the main text. 

\clearpage
\section{Effect of source prompt on the results}
\label{sm:source_prompt}

In the experiments detailed in the main text we paired each image with a source prompt, generated by LLaVA-1.5 and manually refined. However, a source prompt is not required for \ours{} and in fact has little effect on \ours{} results, as detailed in Tab.~\ref{tab:src_prompt}. 
We first tested \ours{}'s sensitivity to the source prompt by generating variations on all the source prompts in the dataset using ChatGPT. We then completely omitted the source prompt and used an empty prompt. Both changes have little effect on the results. These results were obtained using SD3.

\begin{table*}[h]
\centering
\caption{\textbf{Effect of source prompt on the results.} CLIP-T measures adherence to text, while the other four scores measure structure preservation.}
\label{tab:src_prompt}
\begin{tabular}{@{}c|c|cccc@{}}
\toprule
 & CLIP-T $\uparrow$ & CLIP-I $\uparrow$ & LPIPS $\downarrow$ & DINO $\uparrow$ & DreamSim $\downarrow$ \\ \midrule
original source prompt & 0.344 & 0.872 & 0.181 & 0.719 & 0.253 \\
\midrule
different source prompt & 0.343 & 0.872 & 0.181 & 0.713 & 0.254  \\
w/o source prompt & 0.343 & 0.872 & 0.192 & 0.709 & 0.254 \\
\bottomrule
\end{tabular}
\end{table*}

\clearpage
\section{Illustration of the noise-free path between the source and target distributions}
\label{sm:v_delta}

\begin{figure}[h]
    \centering
    \includegraphics[width=\linewidth]{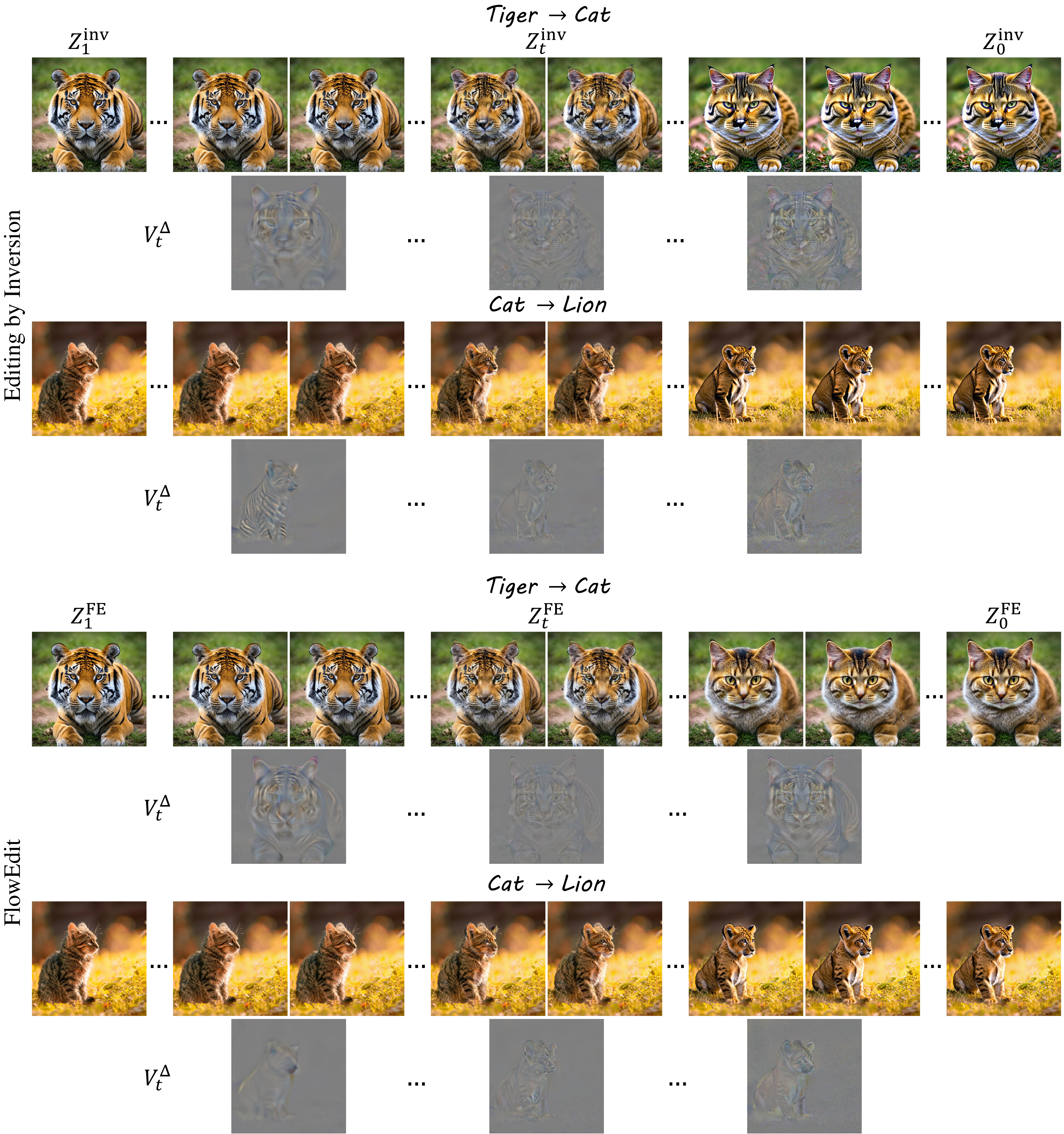}
    \caption{\textbf{Illustration of the noise free path and $V^{\Delta}$.}}
    \label{fig:v_delta}
\end{figure}

As explained in Sec.~\ref{reinterpretation} in the main text, the path defined by \eqref{eq:direct_coupling_ODE_combined} is noise-free, and it constitutes an autoregressive coarse-to-fine evolution from $Z^{\text{src}}_0$ to $Z^{\text{tar}}_0$. Figure \ref{fig:v_delta} illustrates this evolution for both synthetic and real images using editing by inversion (top). This path starts from the source image, $Z_1^{\text{inv}}=Z^{\text{src}}_0$. Moving along this path requires adding the vector $V^{\Delta}_{t}(Z^{\text{src}}_{t},Z^{\text{tar}}_{t})$ to $Z_1^{\text{inv}}$. Illustrations of these vectors along the path are shown beneath the images in Fig.~\ref{fig:v_delta}. Each $V^{\Delta}_{t}(Z^{\text{src}}_{t},Z^{\text{tar}}_{t})$ image is the result of the difference between the two images above it.

It can be seen that for large $t$ values, $V^{\Delta}_{t}(Z^{\text{src}}_{t},Z^{\text{tar}}_{t})$ contains mainly low frequency components. As $t$ decreases, higher-frequency components become increasingly visible. 
This stems from the fact that $V^{\Delta}_{t}(Z^{\text{src}}_{t},Z^{\text{tar}}_{t})$ corresponds to the difference between $V^{\text{tar}}(Z^{\text{tar}}_{t},t)$ and $V^{\text{src}}(Z^{\text{src}}_{t},t)$. 
At large $t$, the noise level is substantial and therefore this vector captures mainly low-frequency components. As $t$ decreases, the vector begins to capture higher-frequency details. 
This process constitutes an autoregressive coarse-to-fine evolution that starts from $Z^{\text{src}}_0$ and ends at $Z^{\text{tar}}_0$, similarly to the evolution of the diffusion/flow process itself~\cite{rissanen2023generativemodellinginverseheat, dieleman2024spectral}.

Figure \ref{fig:v_delta} also illustrates the evolution along the \ours{} path and the $V^{\Delta}_{t}(Z^{\text{src}}_{t},Z^{\text{tar}}_{t})$ along it. These $V^{\Delta}_{t}(Z^{\text{src}}_{t},Z^{\text{tar}}_{t})$ vectors have the same characteristics as in the case of editing by inversion. 

\begin{figure}[h]
    \centering
    \includegraphics[width=\linewidth]{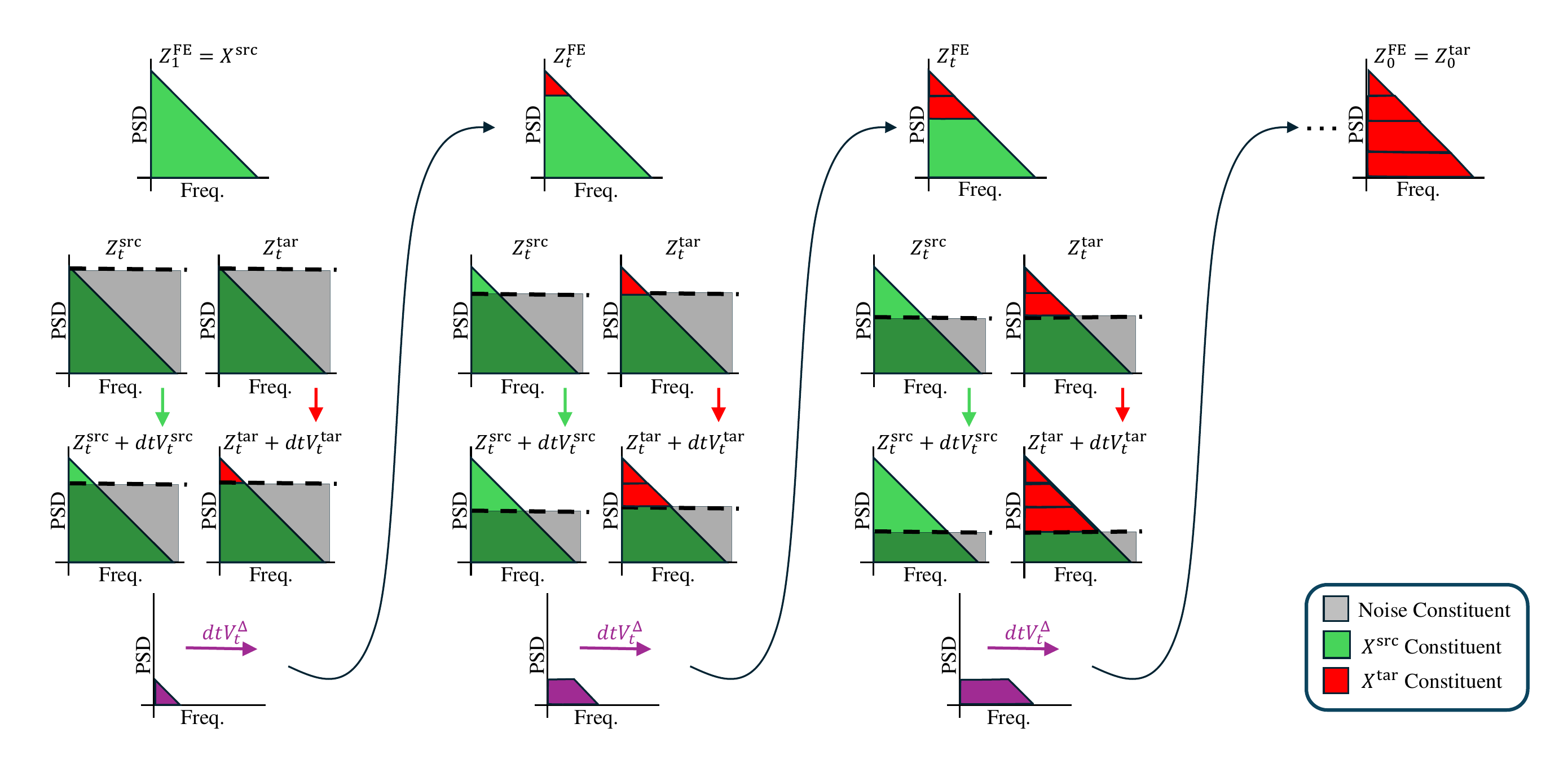}
    \caption{\textbf{Schematic illustration of the spectral behavior of \ours{}'s marginals.} 
    Green and red markings on the graph depict the source and target image's spectral constituents, respectively. The gray markings depict the added white Gaussian noise. 
    The top row illustrates our direct ODE path \eqref{eq:our_ODE}. 
    The second row illustrates our noisy marginals (Sec.~\ref{sec:flowedit}). 
    The third row illustrates the result of a flow step with step-size $dt$ towards the clean distributions. 
    The fourth row illustrates $dtV^{\Delta}_t$. The vector $dtV^{\Delta}_t$ captures the differences between the new source and target frequencies that were ``revealed'' during the denoising step. This $dtV^{\Delta}_t$ is used to drive our ODE process to the next timestep. 
    This is in line with the noise-free, coarse-to-fine behavior we observed in our experiments, and as seen with the image examples in Fig.~\ref{fig:v_delta}. We also provide an empirical evaluation for the illustration in \ref{fig:psd_empiric}.}
    \label{fig:psd_illustration}
\end{figure}

\begin{figure}[h]
    \centering
    \includegraphics[width=\linewidth]{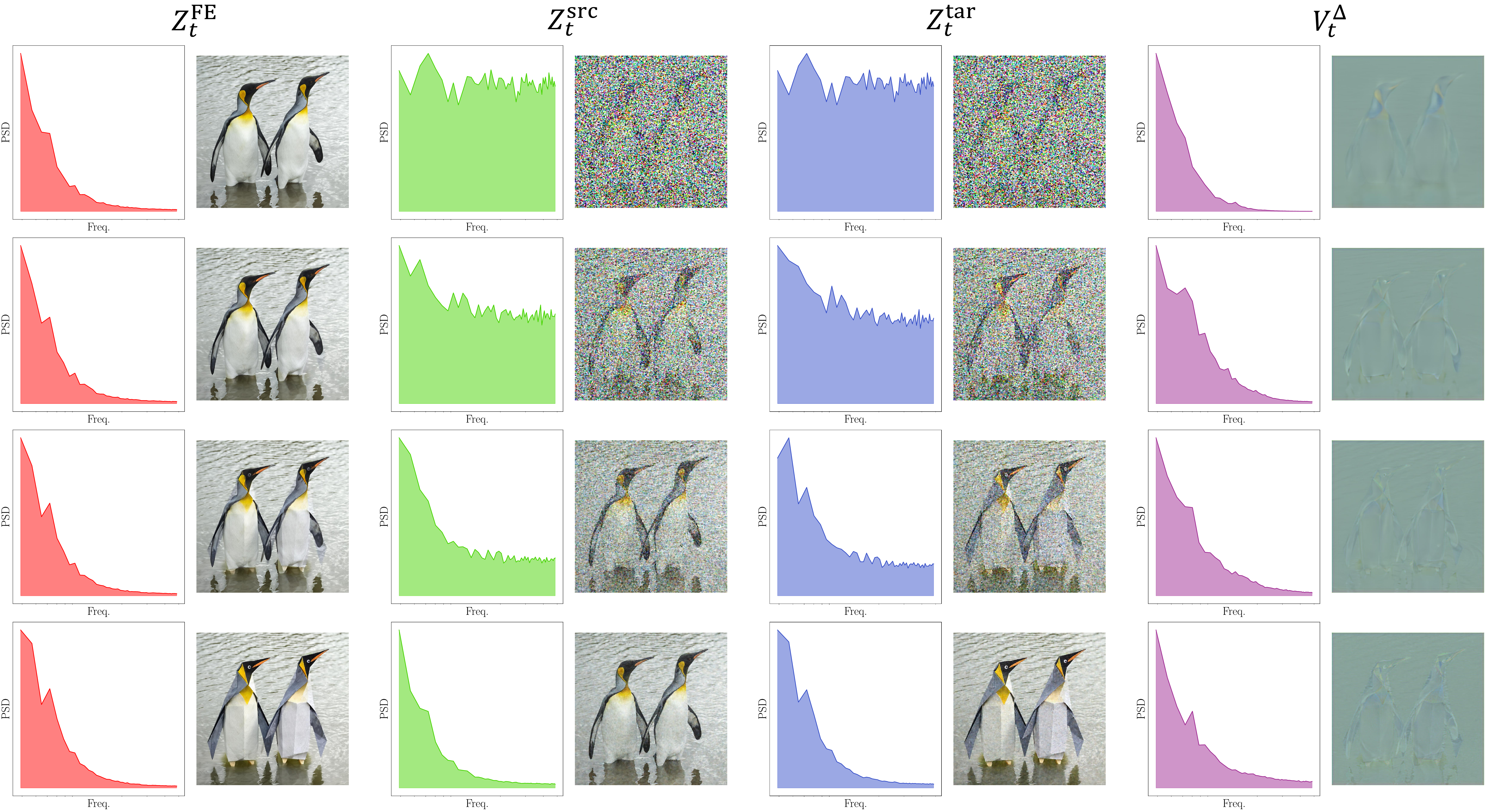}
    \caption{\textbf{Empirical evaluation of  the spectral behavior of \ours{}'s marginals.} 
    The PSD transform \cite{rissanen2023generativemodellinginverseheat} is shown for example images along the \ours{} path (left), the corresponding noisy versions of the source and target images (middle), and the associated velocity fields (right). The PSDs were averaged over four different edits. 
    The example images are the results of one of these edits, corresponding to penguins $\rightarrow$ origami penguins. 
    The first column pair shows the PSD and images of our ODE path $Z^{\text{FE}}_t$ \eqref{eq:our_ODE}, progressing from top to bottom. It can be seen that the spectra of these clean images is nearly the same across all timesteps. 
    The second and third column pairs show the noisy marginals, and as can be seen both $\hat{Z}^{\text{src}}_t$ and $\hat{Z}^{\text{tar}}_t$ are masked with the same amount of noise, and are valid inputs to the trained flow models. 
    Finally, the last column pair represents $V^{\Delta}_t$, which is noise free, and holds more low-frequency edit information in the starting timesteps, and higher frequencies at the later ones. 
    Importantly, these statistics hold for both  (reinterpreted) editing-by-inversion and our method.}
    \label{fig:psd_empiric}
\end{figure}

The autoregressive coarse-to-fine evolution from source to target is also schematically illustrated in Fig.~\ref{fig:psd_illustration} and empirically shown in Fig.~\ref{fig:psd_empiric}. This evolution is illustrated in the frequency domain, using the power spectral density (PSD) transformation, following \citet{rissanen2023generativemodellinginverseheat}.   

\begin{figure}[h]
    \centering
    \includegraphics[width=\linewidth]{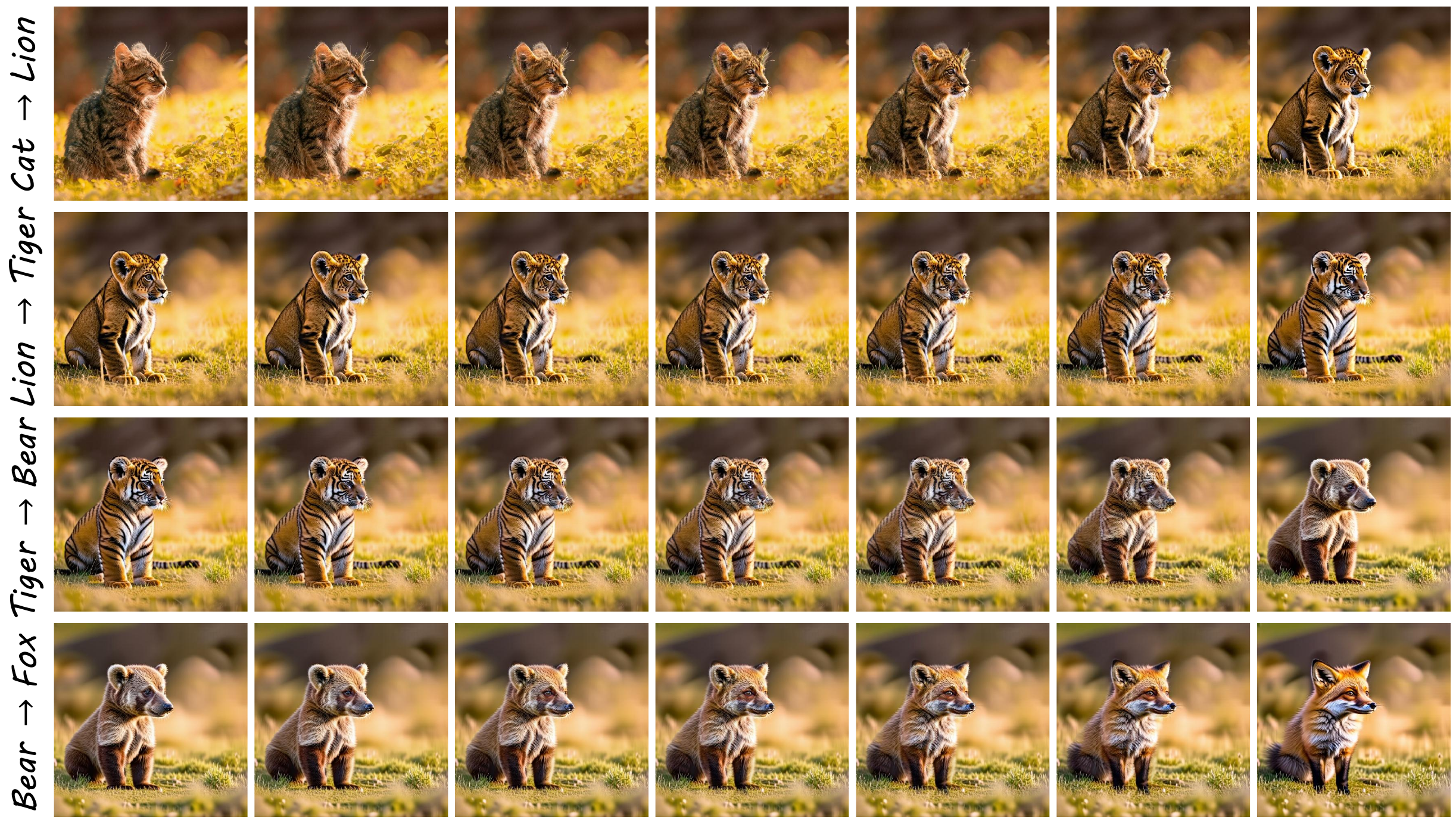}
    \caption{\textbf{Results of editing by inversion along the noise-free path.} The cat image on the top left is used as the input to editing by inversion, where the target prompt is a lion. Then, the lion image is used as input and so forth. It can be seen that the edited images do not fully preserve the structure and fine details of the original image, \textit{e.g} the grass around the cat.}
    \label{fig:editing_interp_ode}
\end{figure}

\begin{figure}[h]
    \centering
    \includegraphics[width=\linewidth]{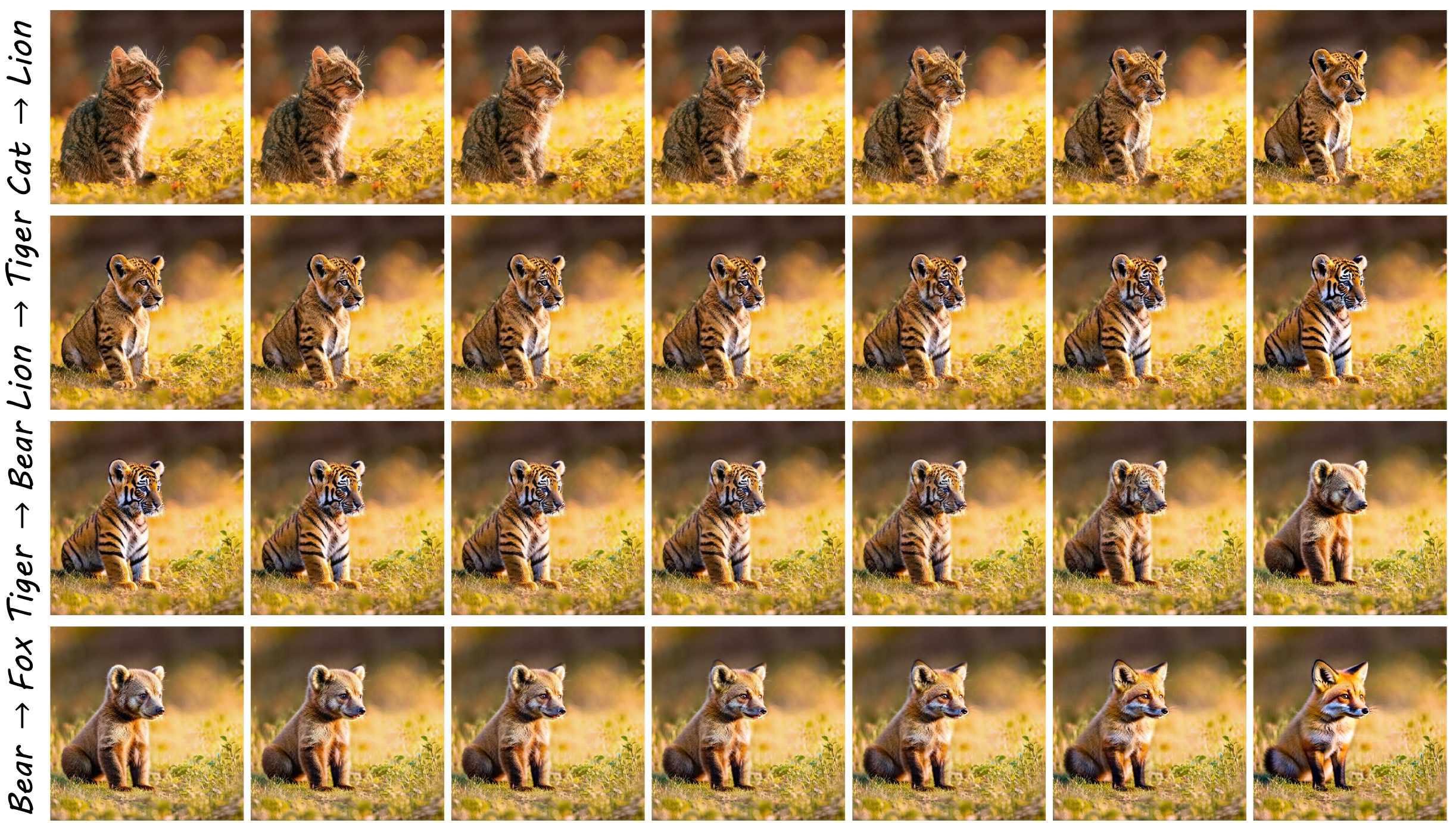}
    \caption{\textbf{Editing results of \ours{} along the noise free path.} The cat image on the top left is used as the input to \ours{}, where the target prompt is a lion. Then, the lion image is used as input and so forth. It can be seen that the edited images preserve the structure and the fine details of the original image, \textit{e.g} the grass around the cat, even after multiple edits.}
    \label{fig:editing_interp}
\end{figure}

By arranging the images $Z_t^{\text{inv/FE}}$ along this path in a video, we can animate the interpolation between the source and target images.
Additionally, by using the resulting target image as the input for subsequent editing steps, we can create a smooth animation that transitions from a source image to multiple edits. This interpolation between the source and target image can be seen in Fig.~\ref{fig:editing_interp_ode} for editing by inversion and in Fig.~\ref{fig:editing_interp} for \ours{}. This noise-free path reveals, for example, how gradually a tiger becomes a bear. Furthermore, the interpolation between a cat image and a fox image, going through lion, tiger and a bear can also be seen at the end of the supplementary video.

\clearpage
\section{Effect of $n_{\text{avg}}$}
\label{sm:practical_navg}
\ours{} operates by evaluating $V^{\Delta}_t(\cdot)$ on multiple realizations of $\hat{Z}^{\text{src}}_t$ and $\hat{Z}^{\text{tar}}_t$ at each $t$. Then, this $V^{\Delta}_t(\cdot)$ is used to drive our ODE \eqref{eq:our_ODE}. This becomes impractical in cases where model evaluations are expensive (SD3/FLUX). Fortunately, averaging also occurs across timesteps when the noises $\{N_t\}$ are chosen to be independent across timesteps. This means that with large enough $T$, we can use a smaller $n_{\text{avg}}$, reducing expensive model evaluations. Specifically, the default values of $T$ for SD3 and FLUX are high enough for our method to perform well with $n_{\text{avg}}=1$, relying purely on averaging across $t$.

\begin{figure}[h]
     \centering
    \includegraphics[width=0.45\linewidth]{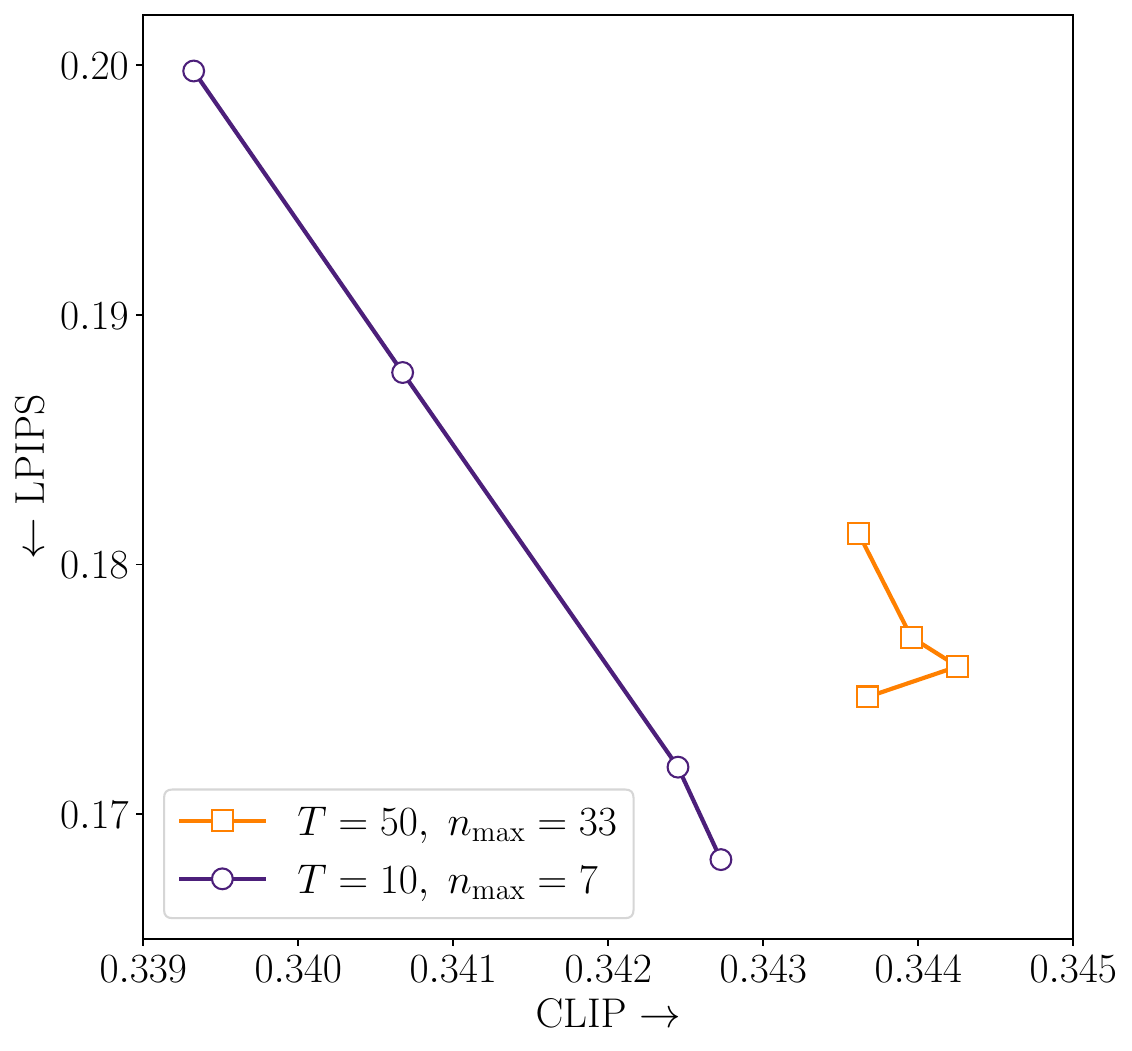}
    \caption{\textbf{CLIP vs. LPIPS for different values of $n_{\text{avg}}$.} From top to bottom, the markers correspond to $n_{\text{avg}}$ of $1,\; 3,\; 5,\; 10$.}
    \label{fig:diff_n_avg}
\end{figure}

Figure \ref{fig:diff_n_avg} illustrates this effect. As the value of $n_{\text{avg}}$ increases, the LPIPS distance decreases. We used SD3 and $n_{\text{avg}}$ values of $1,\; 3,\; 5,\; 10$. 
For a large number of discretization and editing steps, \ie $T=50,\;n_{\text{max}}=33$ (orange curve), the number of averaging steps has little effect on the results, as averaging already occurs between timesteps. Hence, increasing $n_{\text{avg}}$ only slightly improves the LPIPS score and has a negligible effect on the CLIP score (note that the horizontal axis ticks spacing is $0.001$). These hyperparameters, $T=50,\;n_{\text{max}}=33$, with $n_{\text{avg}}=1$, are the hyperparameters used in our method.
However, for a small number of discretization and editing steps, \ie $T=10,\;n_{\text{max}}=7$ (purple curve), increasing $n_{\text{avg}}$ has a substantial effect. It reduces the LPIPS distance by $\sim0.3$ and increases the CLIP score by $0.035$.
In both experiments, the CFG scales are the same as those described in the main text.

\clearpage
\section{Further discussion on the relation to optimization based methods}
\label{sm:dds}

As detailed in the main text, it is unnatural to view \ours{} as an optimization based method, like DDS or PDS, because (i) it does not choose timesteps $t$ at random but rather following a monotonically decreasing schedule $\{t_i\}^{n_{\max}}_{i=0}$, and (ii) it must use a learning rate of exactly $dt=t_{i-1}-t_{i}$ at iteration $i$. In Sec.~\ref{sec:stepsize} below, we show that even a slight change of the step size in \ours{} leads to rapid deterioration of output quality. This suggests that it is more natural to view \ours{} as solving an ODE rather than an optimization problem.

Point (i) above is a major distinction between \ours{} and DDS/DPS, as it guarantees that the model is always fed with inputs from the distribution on which it has been trained. When sampling timesteps at random, as in DDS and DPS, there is a possibility of drawing a small timestep $t$ at the beginning of the process, when the image has not yet been modified. This leads to out-of-distribution inputs to the model. For example, in the cat$\rightarrow$lion edit of Fig.~\ref{fig:editing_interp}, if DDS draws a small $t$ at the beginning of the process, then the model is fed with a cat image that is only slightly noisy (where the cat is clearly visible), and is tasked with denoising it conditioned on the text ``lion''. Clearly, a slightly noisy cat image is out-of-distribution under the condition ``lion''. 

However, it turns out that there is an even more fundamental issue with the optimization viewpoint. Specifically, we claim that the optimization viewpoint is not fully justified even for DDS, as the loss that it attempts to minimize does not actually decrease throughout the DDS iterations. We thoroughly evaluate this in the next subsection. 

\subsection{Delta denoising loss}

\begin{figure}[h]
    \centering
    \includegraphics[width=\linewidth]{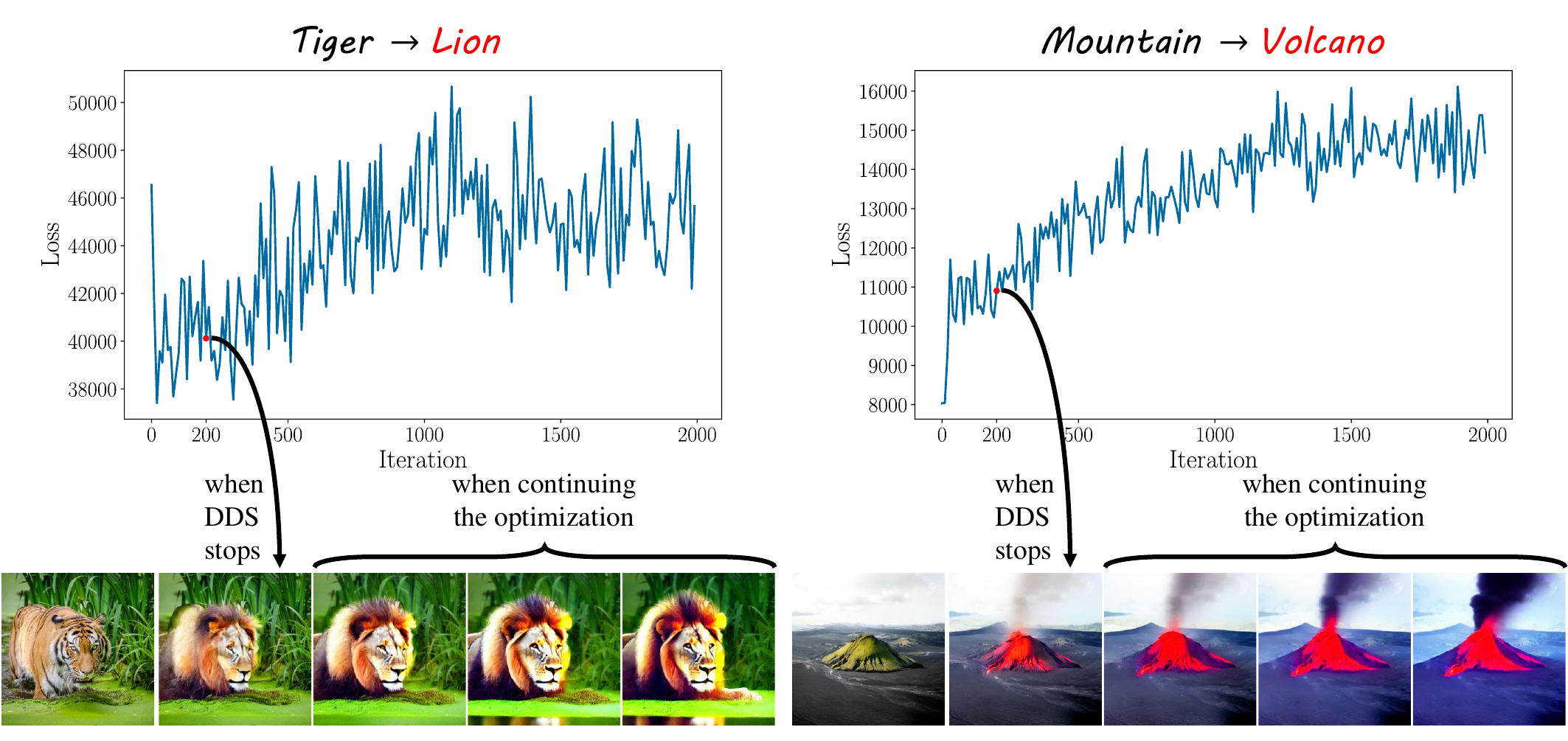}
    \caption{\textbf{DDS optimization process.}}
    \label{fig:dds}
\end{figure}

The DDS method was presented as an iterative approach for minimizing the delta denoising (DD) loss.
The DD loss for matched and unmatched image-text embedding pairs $\mathbf{z},y,\mathbf{\hat{z}},\hat{y}$ respectively, is defined as 
\begin{equation}\label{eq:ddloss}
    \mathcal{L}_{\text{DD}}(\phi,\mathbf{z},y,\mathbf{\hat{z}},\hat{y},\epsilon)= \int_0^1\| \epsilon^\omega(\mathbf{z}_t,y,t)-\epsilon^\omega(\mathbf{\hat{z}}_t,\hat{y},t)\|^2 dt, 
\end{equation}
where $t$ is the diffusion timestep and $\epsilon^\omega(\cdot)$ is the trained (noise-predicting) model with guidance scale $\omega$~\cite{hertz2023delta}. The DDS iterations constitute an approximation for a stochastic gradient descent process. However, as we illustrate in Fig.~\ref{fig:dds}, this approximation is quite poor.
Specifically, as can be seen for two editing examples, the DDS iterations do not decrease the DD loss. They rather tend to increase it. Furthermore, as illustrated in Fig.~\ref{fig:dds}, if the optimization is allowed to continue, the editing results deteriorate. 

These experiments were carried out using the official DDS implementation\footnote{\url{https://github.com/google/prompt-to-prompt/blob/main/DDS_zeroshot.ipynb}}, with SD 1.5~\cite{rombach2022high}. 
We allowed the DDS optimization process to continue beyond the default $200$ iterations used in the official implementation by running the optimization process an additional $1800$ iteration, to a total of $2000$ iterations.
Every $100$ iterations we calculated the DD loss (\eqref{eq:ddloss}) for $19$ timesteps between $50$ and $950$ with spacing of $50$. We then summed the loss for all these timesteps to obtain the loss vs.~iteration graphs in Fig.~\ref{fig:dds}.  

We do not compare \ours{} with DDS results directly, as the comparison will not be fair due to the different backbone models, but we rather shed light on this strange behavior of the DDS optimization process.  

\subsection{\ours{} with different step sizes}
\begin{figure}[h]
    \centering
    \includegraphics[width=0.9\linewidth]{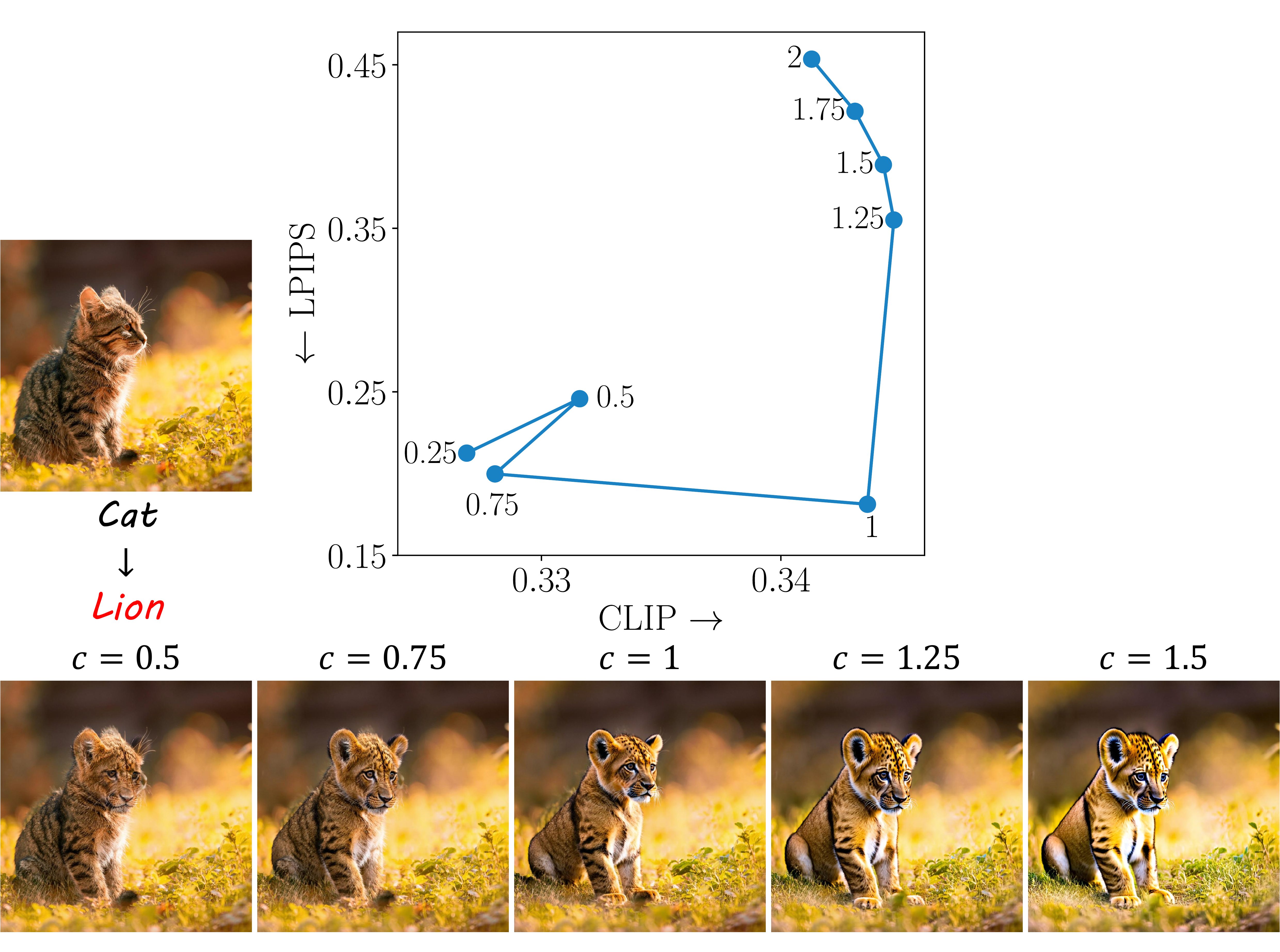}
    \caption{\textbf{\ours{} results with a scaled step size.} The scale parameter $c$ that multiplies the step-size is indicated next to each point.}
    \label{fig:eta}
\end{figure}

When using \ours{} with a step size that is either smaller or larger than $dt=t_{i-1}-t_{i}$, the results deteriorate, as illustrated in Fig.~\ref{fig:eta}, qualitatively and quantitatively. 
We evaluate \ours{} using SD3 with step-size, $dt$, scaled by a factor of $c$. Only for $c=1$ we get a favorable balance between the CLIP and LPIPS metrics, \ie a high CLIP score and a low LPIPS score. 
Furthermore, we see that even slight deviations from $c=1$ lead to a significant drop in performance. 
This can also be clearly seen in the qualitative example, where the edited result adheres to the text prompt and is free of artifacts only for $c=1$. 

This behavior indicates that \ours{} cannot be considered as a gradient descent (GD) optimization over a loss function. In GD, step sizes with similar values yield similar results, and the optimization dynamics are generally smooth, without abrupt changes. This is different from \ours{} results illustrated in Fig.~\ref{fig:eta}, where a big difference in the results occurs with $c$ values around $1$. 
In fact, as \ours{} solves an ODE that traverses a path between the source distribution and the target distribution, the step sizes along this path need to sum to $1$. Arbitrarily scaling the step-size by some constant violates this requirement and therefore leads to deteriorated results.     

\subsection{Mathematical discrepancies between the \ours{}, DDS, PDS update steps}
\label{sec:stepsize}
Regardless of the key differences in the ODE vs.~optimization viewpoints, here we focus on the difference between the update steps themselves, translated to a common diffusion denoising formulation. To do so, we rewrite our ODE step $V^\Delta_t$ in terms of differences in noise predictions. Specifically, assuming $X_t=(1-t)X_0+t\epsilon_t$, we will express the velocity field, which is given by $V(x_t,t)=\mathbb{E}[\epsilon_t-X_0|X_t=x_t]$ (see \cite{liu2023flow}) in terms of the noise prediction $\hat{\epsilon}(x_t,t)\triangleq\mathbb{E}[\epsilon|x_t=x_t]$.
We have
\begin{align}
    V(x_t,t)&= \mathbb{E}[\epsilon-X_0|X_t=x_t] \nonumber \\
    &= \frac{1}{1-t}\mathbb{E}[(1-t)\epsilon-(1-t)X_0|X_t=x_t] \nonumber \\
    &= \frac{1}{1-t}\mathbb{E}[\epsilon-((1-t)X_0+t\epsilon)|X_t=x_t] \nonumber \\
    &= \frac{1}{1-t}(\mathbb{E}[\epsilon|X_t=x_t]-x_t) \nonumber \\
    &= \frac{1}{1-t}(\hat{\epsilon}(x_t,t)-x_t).
\end{align}
Therefore,
\begin{align}
        V^{\Delta}_{t}(\hat{Z}^{\text{src}}_{t},\hat{Z}^{\text{tar}}_{t}) &= V^{\text{tar}}(\hat{Z}^{\text{tar}}_{t},t)-V^{\text{src}}(\hat{Z}^{\text{src}}_{t},t) \nonumber\\
        &= \frac{1}{1-t}\left(\hat{\epsilon}(\hat{Z}^{\text{tar}}_{t},t)-\hat{Z}^\text{tar}_t\right)-\frac{1}{1-t}\left(\hat{\epsilon}(\hat{Z}^{\text{src}}_{t},t)-\hat{Z}^\text{src}_t\right) \nonumber \\
        &= \frac{1}{1-t}\left(\left(\hat{\epsilon}(\hat{Z}^{\text{tar}}_{t},t)-\hat{\epsilon}(\hat{Z}^{\text{src}}_{t},t)\right)-\left(\hat{Z}^\text{tar}_t-\hat{Z}^\text{src}_t\right)\right) \nonumber \\
        &= \frac{1}{1-t}\left(\left(\hat{\epsilon}(\hat{Z}^{\text{tar}}_{t},t)-\hat{\epsilon}(\hat{Z}^{\text{src}}_{t},t)\right)-\left(Z^{\text{FE}}_t-tX^{\text{src}}_0+t\epsilon_t-(1-t)X^{\text{src}}_0-t\epsilon_t\right)\right) \nonumber \\
        &= \frac{1}{1-t}\left(\left(\hat{\epsilon}(\hat{Z}^{\text{tar}}_{t},t)-\hat{\epsilon}(\hat{Z}^{\text{src}}_{t},t)\right)-\left(Z^{\text{FE}}_t-X^{\text{src}}_0\right)\right),
\end{align}
where the fourth transition comes from our construction of $\hat{Z}^{\text{src}}_{t} =(1-t)Z_0^{\text{src}}+t\epsilon_t$ and $\hat{Z}^{\text{tar}}_t=Z^{\text{FE}}_t+\hat{Z}^{\text{src}}_t-Z^{\text{src}}_0$ which we discussed in the main text. 

As we discussed earlier, the update step in DDS is $\eta(t)(\hat{\epsilon}(Z^{\text{tar}}_t,t)-\hat{\epsilon}(Z^{\text{src}}_t,t)$, and is different from our update step, since it only relies on the differences between the noise predictions. 

Regarding PDS\cite{koo2024posterior}, the proposed update step in Eq.~(14) in their paper is of the form
\begin{align}
    \nabla\mathcal{L}_{\text{PDS}}=\psi(t)(Z^{\text{PDS}}_t -X^{\text{src}}_0)+\chi(t)(\hat{\epsilon}(Z^{\text{tar}}_t,t)-\hat{\epsilon}(Z^{\text{src}}_t,t),
\end{align}
where $Z^{\text{PDS}}_t$ is the variable that is being optimized, and $\psi(t),\chi(t)$ are diffusion-dependent coefficients. Note that their update step can only coincide with ours (up to a scalar multiplier) if $\forall t:\; \psi(t)=-\chi(t)$, which is not the case in their paper. 
Moreover, the authors of PDS state that their optimization scheme works when $t$ is selected at random, and if chosen sequentially (mimicking \ours{}) the gradients zero out (see the paragraph after Eq.~(29) in PSD supplementary material). This further exacerbates the difference between the methods.

We conclude that even though DDS and PDS  appear similar, when reformulating our update step in diffusion terms, the mathematical differences become clearer. This is added to the fact that both DDS and PDS attempt to solve an optimization problem (minimize differences between two vectors), which as we showed earlier is not fully justified.

\clearpage
\section{Additional details about the Cats-Dogs experiment}
\label{sm:cats_dogs}

\begin{figure}[h]
    \centering
    \includegraphics[width=0.7\linewidth]{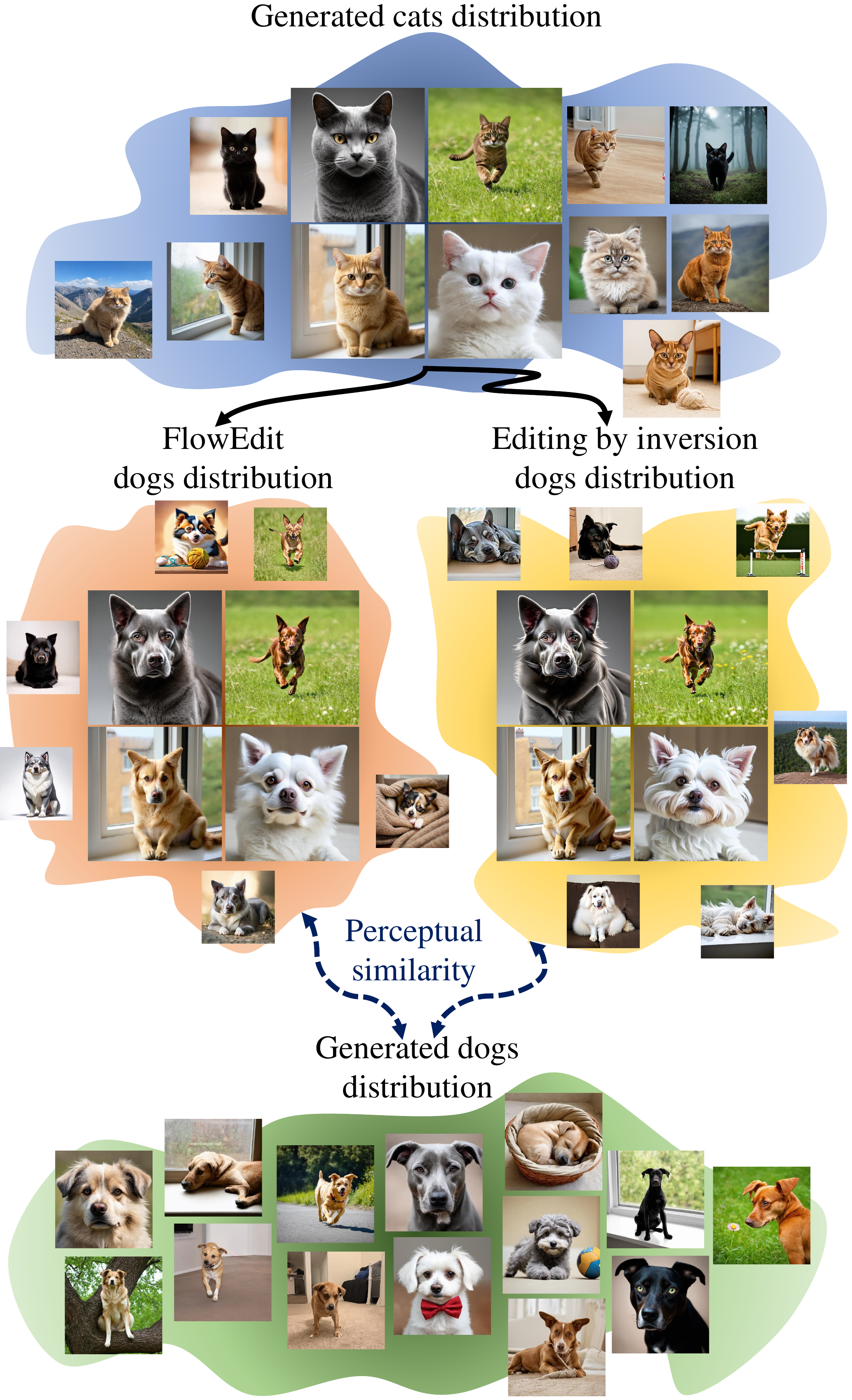}
    \caption{\textbf{Illustration of the Cats-Dogs experiment.}}
    \label{fig:cat_dogs_sm}
\end{figure}

In Sec.~\ref{sec:cats_dogs_exp} we described experiments evaluating the reduced transport cost of \ours{} compared to editing by inversion starting with generating a synthetic dataset of cat images. To generate the 1000 cat images we used variations of the prompt ``a photo of cat'' generated by Llama3~\citep{dubey2024llama3} 8B instruction model. 
Specifically, we requested 1000 source prompts describing cat images by providing the instruction: ``Output 1000 possible prompts to a text-to-image model. These prompts should be a variation of `a photo of a cat.', by adding descriptions and adjectives.'' 
We used these 1000 prompts as input to SD3 and generated with them 1000 cats images. Examples of these generations can be seen in the upper part of Fig.~\ref{fig:cat_dogs_sm}, covering the blue shape. 

To create corresponding edited dog images, we applied both editing by inversion and \ours{} using target prompts identical to the source prompt, except for replacing the word ``cat'' with ``dog''. 
Examples of these editing results can be seen in the middle of Fig.~\ref{fig:cat_dogs_sm}, covering the orange shape for \ours{} results and covering the yellow shape for editing by inversion. 
Specifically, the four dog images in the middle of each shape are the results of editing the four cat images in the middle of the blue shape. 
It can be seen that \ours{} editing are better when compare to editing by inversion results. 
We also calculate the transport cost between the cats distribution and both dogs distributions. We did it by calculating the average squared distance between SD3 latent representation of the cat images and their paired dog images, for both edits. 
We also calculated LPIPS between the original cat images and their paired dog images. 
As presented in Sec~\ref{sec:cats_dogs_exp}, in both metrics, \ours{} achieved lower transport cost, compared to editing by inversion.       

In addition, we generated 1000 dog images using SD3 and the same text prompts used to generate the cat images, but with ``cat'' replaced by ``dogs''. 
Examples of these generations can be seen in the bottom part of Fig.~\ref{fig:cat_dogs_sm}, covering the green shape. 
To asses the alignment between the edited dogs distribution and the generated dogs distributions we used FID and KID scores. As shown in Sec~\ref{sec:cats_dogs_exp}, 
\ours{} achieved lower FID and lower KID scores, indicating our ability to produce images from the target distribution.

